\documentclass[lettersize,journal]{IEEEtran}
\usepackage{amsmath,amsfonts}
\usepackage{array}
\usepackage{textcomp}
\usepackage{stfloats}
\usepackage{url}
\usepackage{verbatim}
\usepackage{graphicx}
\usepackage{pifont}
\usepackage{booktabs}
\usepackage{subcaption}
\usepackage{algorithm}
\usepackage{amssymb}
\usepackage{algpseudocode}
\usepackage[algo2e]{algorithm2e}
\usepackage{color}
\usepackage{diagbox}
\usepackage{bbding}

\SetKwInput{KwInput}{Input}                
\SetKwInput{KwOutput}{Output}              
\usepackage{textcomp}
\usepackage{pgfplots}
\usepackage{multirow}
\usepackage{cite}
\usetikzlibrary{
  pgfplots.groupplots,
  matrix
}
\usepackage{siunitx}

\def\BibTeX{{\rm B\kern-.05em{\sc i\kern-.025em b}\kern-.08em
    T\kern-.1667em\lower.7ex\hbox{E}\kern-.125emX}}
\usepackage{balance}

\title{Overcoming Support Dilution for\\ Robust Few-shot Semantic Segmentation}

\author{%
  Weiliang Tang\textsuperscript{*} \\
  The Chinese University of Hong Kong\\
  \texttt{wltang21@cse.cuhk.edu.hk} \\

  Biqi Yang\textsuperscript{*} \\
  The Chinese University of Hong Kong \\
  \texttt{bqyang@cse.cuhk.edu.hk} \\
  
  Pheng-Ann Heng \\
  The Chinese University of Hong Kong\\
  \texttt{pheng@cse.cuhk.edu.hk} \\
  
  Yunhui Liu \\
  The Chinese University of Hong Kong\\
  \texttt{yhliu@mae.cuhk.edu.hk}
  
  Chi-Wing Fu \\
  Department of CSE and SHIAE \\
  The Chinese University of Hong Kong \\
  \texttt{cwfu@cse.cuhk.edu.hk} \\
  \renewcommand\footnotemark{}

  \thanks{
  \textsuperscript{*} Equal contributions to the works. This work is supported by the InnoHK Clusters of the Hong Kong SAR Government via the Hong Kong Centre for Logistics Robotics}
}

\begin{document}
\maketitle

\begin{abstract}
Few-shot Semantic Segmentation (FSS) is a challenging task that utilizes limited support images to segment associated unseen objects in query images. 
%
%
%
However, recent FSS methods are observed to perform worse, when enlarging the number of shots.
As the support set enlarges, existing FSS networks struggle to concentrate on the high-contributed supports and could easily be overwhelmed by the low-contributed supports that could severely impair the mask predictions.
In this work, we study this challenging issue, called support dilution, our goal is to recognize, select, preserve, and enhance those high-contributed supports in the 
raw support pool. 
%
%
%
Technically, our method contains three novel parts.
First, we propose a contribution index, to quantitatively 
estimate if a high-contributed support dilutes.
Second, we develop the Symmetric Correlation (SC) module to preserve and enhance the high-contributed support features, minimizing the distraction by the low-contributed features.
Third, we design the Support Image Pruning operation, to retrieve a compact and high-quality subset by discarding low-contributed supports. 
We conduct extensive experiments on two FSS benchmarks, COCO-20$^i$ and PASCAL-5$^i$, the segmentation results demonstrate the compelling performance of our solution over state-of-the-art FSS approaches. 
Besides, we apply our solution for online segmentation and real-world segmentation, convincing segmentation results showing the practical ability of our work for real-world demonstrations. 
\end{abstract}


\section{Introduction}
Semantic segmentation is a fundamental vision task, supporting a wide range of real-world applications, such as robotics manipulation, medical image analysis, and autonomous driving~\cite{wong2021manipulation, schwarz2018rgb, lin2018visual, khan2021deep, yang2021artificial, huang2020deep, xiao2023baseg, chen2017importance}. 
Though deep-learning-based approaches~\cite{yang2018denseaspp, yang2021artificial, huang2019ccnet, ronneberger2015u, zhao2017pyramid, lin2017refinenet} demonstrate remarkable performance, they generally require a large volume of annotated data to enable model learning. 
Large performance drops are, however, often observed when handling novel classes that are not in the training data.

%
%
To tackle this generalization problem, Few-shot Semantic Segmentation (FSS) methods have been proposed~\cite{min2021hypercorrelation, hu2019attention, lang2023base, iqbal2022msanet, peng2023hierarchical, shi2022dense, xu2023self, li2021adaptive}.
The idea is to utilize a small amount of annotated samples (supports), and segment test samples (queries) of the novel class by the learned support-query correlations.
Typically, for $N$-shot FSS, the support set contains $N$ exemplars, which are manually prepared, and $N$ is usually only 1, 3, or 5 for evaluation purposes in common experimental inference.

\begin{figure}
    \centering
    \includegraphics[width=0.5\textwidth]{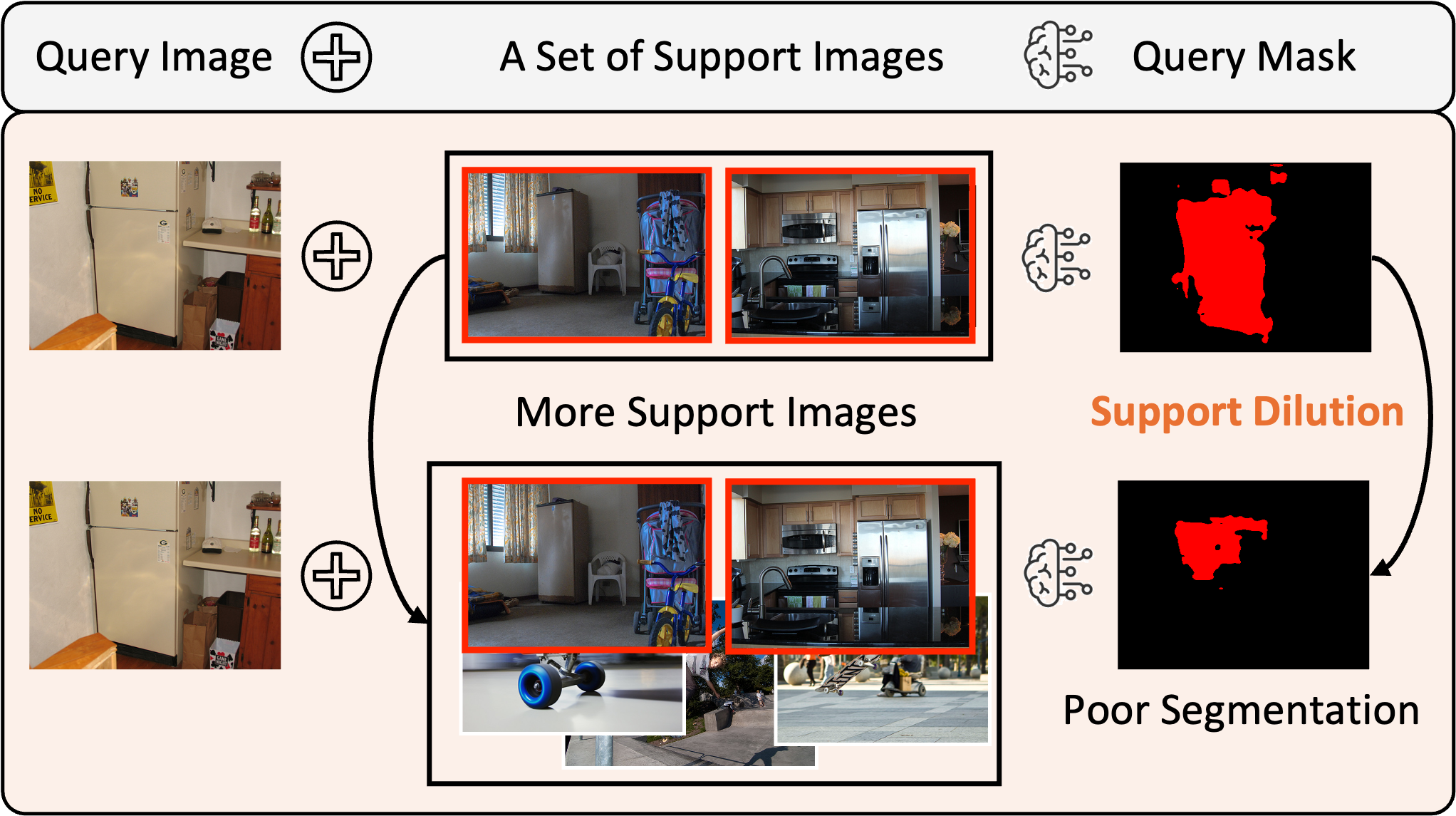}
    \caption{When the number of supports gets larger, SOTA FSS method DCAMA~\cite{shi2022dense} (ECCV'22) cannot concentrate on the high-contributed supports and are distracted by the low-contributed supports. In this figure, we increase $N$ from 2 to 30, in the 30-support set, we omit some supports for briefness.}
    \label{fig: teaser}
\end{figure}
%
To improve the segmentation performance, an intuitive solution is to feed the FSS networks with more support images, e.g., 5-shot FSS often outperforms 1-shot FSS. 
However, an 
empirical observation is that a bold growth of $N$ may not guarantee a consistent performance gain.  
See Fig.~\ref{fig: teaser}, given a larger support set ($N$ from 2 to 30), the state-of-the-art (SOTA) FSS approach DCAMA~\cite{shi2022dense} turns out under-segmentation for the refrigerator object. 
A more comprehensive experiment will be presented in Sec.\ref{sec:preliminary studies}, in which the same phenomenon is observed.
That is, when $N$ continuously gets larger, the useful support information is gradually diluted by the noise and irrelevant information. Such an issue, we coin as \textbf{\textit{support dilution}}, leads to the performance drop.
%

%
Support dilution is a critical challenge to deploying FSS methods in real-world applications. 
%
In today's information age, we collect vast and inexhaustible support images by various data-mining techniques (e.g., via image searching API) from different data sources, usually without filtering, alignment, or refinement.
%
%
In real situations, we will obtain a large and possibly dirty support pool, in which some representable supports can positively guide the segmentation, while most others contain little usable information.
%
We define the former supports as high-contributed supports (bounded red in Fig.~\ref{fig: teaser}) and the latter ones as low-contributed supports.
For prior FSS methods, we have to pay a tedious {\em manual\/} effort to pick the high-contributed samples; otherwise, the segmentation masks will be severely impaired by the low-contributed supports. In other words, we either sacrifice the method efficiency, or sacrifice the method reliability.
To deal with the support dilution problem, in this work, we propose a novel
%
solution to automatically recognize, accurately select, then preserve and enhance the high-contributed supports out of the massive yet noisy information pool, for generating precise segmentation masks.
Our solution includes three parts.
\textbf{Contribution Index.} 
Support dilution means the high-contributed supports contribute less to the query, while the low-contributed supports show much effort and dilute the high-contributed ones.
%
%
We design an index to quantitatively estimate 
the true contribution of each support, in which the values are determined by 
correlation (i.e., attention) weight between its encoded feature and the query feature. 
%
%
%
%
%
%
%
%
Importantly, A good FSS framework should consistently preserve the contribution values of the high-contributed supports and suppress the contribution values of the low-contributed supports. This lead to our second design.
%

%
\textbf{Symmetric Correlation.}
Previous FSS correlation modules~\cite{iqbal2022msanet, shi2022dense, xu2023self, hu2019attention} can extract rich information from high-contributed supports, but they are also sensitive to (or distracted by) the low-contributed supports.
%
%
We concentrate on an interesting case, where we use the query itself as one of the support inputs, and we call it upper-bound support (i.e., the highest-contributed support, which results in the upper-bound segmentation performance). 
We find that the contribution value of our upper-bound support decreases as $N$ gets larger, indicating that existing correlation modules lack robustness against heavy information noise, the high-contributed supports dominate less and the low-contributed supports become more influential.
To tackle this problem, we propose Symmetric Correlation (SC), which can preserve and enhance the high-contributed support features in support-query correlation learning. 
The key idea is to ensure the correlation score to reach the maximum when and only when we input an identical support-query pair.
With this constraint, our upper-bound support can permanently obtain the largest contribution value, and other high-contributed support features are simultaneously consolidated depending on their visual similarity with the query.
%
%
%

%
\textbf{Support Image Pruning.}
%
%
In real-world applications, the numbers of high-contributed and low-contributed supports can be highly imbalanced. 
Overwhelming low-contributed features can sum up to a large score, thus impairing the ability of SC. 
Besides, when $N$ is large, we have to calculate correlation scores for all those low-contributed supports and pay redundant SC computation efforts.
%
%
%
A direct idea is to cut down the support set before the correlation calculation, such that SC can ignore the irrelevant and noisy information and only focus on those relevant supports.
Hence, we propose the Support Image Pruning operation. 
After this automatic pruning operation, we can then sufficiently and adaptively yield the ability of SC on a compact and high-quality subset.
Pruning is a subset retrieval task, to solve this task, 
we design a contribution-guided greedy algorithm.
Given the original large set, we push the items (i.e., supports) into a new queue (i.e., subset) one by one. 
In each iteration, we retrieve the support that can maximize the current overall contribution, at last we produce a sub-optimal result with low time complexity. 

In this paper, our contributions can be summarized as:
\begin{itemize}
    \item We revisit the FSS setting and explore the support dilution problem (Sec.~\ref{sec:preliminary studies}), which is critical in the real world but has long been ignored. Given more support images, existing FSS methods struggle to focus on the high-contributed supports and are easily distracted the low-contributed ones. We propose an effective solution to tackle support dilution.
    %
\item Technically, our solution has three parts (Sec~\ref{sec: method}). First, we design a contribution index, which quantitatively estimates the true contribution for each support (Sec.~\ref{method CF}). Second, we propose Symmetric Correlation (SC), which helps to preserve the high-contributed supports, while avoiding negative distractions of the low-contributed supports (Sec.~\ref{method SC}). Last, we propose the Support Image Pruning operation, where we retrieve a smaller subset for information purification (Sec.~\ref{method SIP}).
    \item We conduct extensive experiments. Quantitative and qualitative results on COCO-20$^i$ and PASCAL-5$^i$ show the superiority of our method over SOTAs. We also deploy our method for online FSS and real-world FSS to manifest its practicality. Details can be found in Sec.~\ref{exp}.
\end{itemize}
\begin{figure*}
    \centering
    \includegraphics[width=\textwidth]{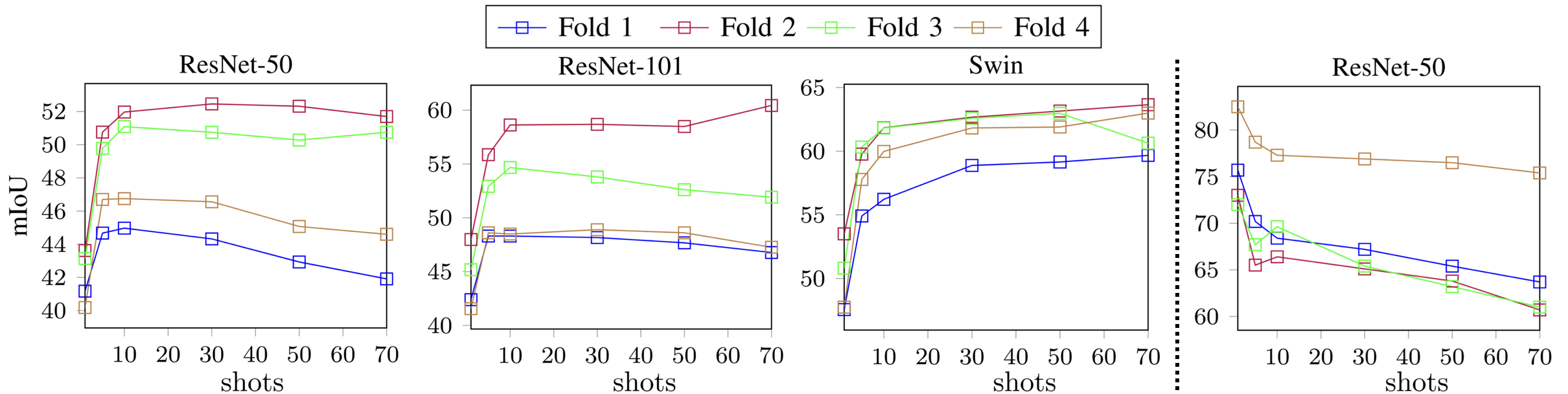}
    \caption{Left: the performance of DCAMA cannot gain consistent improvements when the number of shots $N$ gets larger. Right: when mixing more noisy information, DCAMA ResNet-50 cannot protect the upper-bound support from dilution and we observe a drastic performance drop.}
    \label{fig: preliminary_exp_result}
\end{figure*}

\section{Related Work}
\subsection{Semantic Segmentation}
Semantic segmentation (SS) is one of the fundamental computer vision tasks, aiming at segmenting put every pixel of the image into pre-defined categories. Fully convolution neural networks (FCNs) are the cornerstone that empower the network with the visual ability. Impressive progress has been made with it. Since then, a variety of techniques have been developed to improve the SS performance. The main motivation of these techniques can be split into branches: (1) Broaden the receptive-field~\cite{yang2018denseaspp, yu2015multi, huang2019ccnet} and (2) Harness multi-level features~\cite{zhao2017pyramid, ronneberger2015u, lin2017refinenet}.
\subsection{Few-shot Learning}
Deep neural network (DNN) suffers from over-fitting problems and is poor to generalize to unseen categories when the number of training data is scarce. Few-shot learning (FSL) is introduced to tackle the issues. In the FSL task, there are seen categories with adequate training samples and unseen categories with only limited training samples (e.g., 1, 5). The prevailing methods can be split into three branches: (1) Optimization-based~\cite{finn2017model, jamal2019task, ravi2016optimization, wang2003frustratingly, vinyals2016matching}; (2) Augmentation-based~\cite{chen2019image, chen2019image2, ghiasi2021simple, feng2023diverse, trabucco2023effective}; and (3) Metric-based~\cite{li2019finding, snell2017prototypical, sung2018learning, wu2021learning, vinyals2016matching}. The optimization-based methods propose a better training paradigm or better optimization target to tackle with over-fitting and the data bias caused by imbalance training samples. The augmentation-based methods improve the DNN's generalization and prevent over-fitting by introducing various data augmentation. It can be either hand-crafted~\cite{ghiasi2021simple, chen2019image} or generated with models~\cite{trabucco2023effective, feng2023diverse}. The metric-based methods develop a general metric space to measure the similarity between the test and training samples. Prototypical network~\cite{snell2017prototypical} proposes the concept of prototypes that model the common characteristics of a class. Following the idea, \cite{dong2018few} calculates a finer class prototype by applying masked average pooling. \cite{wang2019panet} adds extra regularization to better align prototypes and the test sample. Instead of calculating feature distances, \cite{sung2018learning, zhang2019canet} learns networks to predict the sample similarity with prototypes. 
\subsection{Few-shot Semantic Segmentation}
Few-shot semantic segmentation (FSS) classifies each pixel in the test image (query image) with unseen class provided with only a few of its images (support images). The FSL work evolves in the direction of utilizing finer-and-finer-grained support information (e.g., from class-level, part-level to pixel-level information). \cite{shaban2017one, zhang2020sg} are the early work that harnesses class-level or instance-level prototypes. PFENet~\cite{tian2020prior} introduces a training-free prior mask that calculates the similarity in the high-level feature to provide mask prior. ASGNet~\cite{li2021adaptive} learns finer-grained correlation by learning sub-instance prototypes.  HSNet~\cite{min2021hypercorrelation} generates a 4D correlation map by calculating pairwise feature similarity between the query and the support image. \cite{hu2019attention} fuses support features with correlation operation in multi-scale to provide segmentation guidance. \cite{lang2023base} predicts region-wise correlation between the query and the seen categories to suppress the contribution of irrelevant classes. SCCAN~\cite{xu2023self} learns query-support mutual correlation with the Swin transformer~\cite{liu2021swin}. It also proposes self-calibrated cross-attention module to resolve the misalignment between the query background and the support foreground. MSANet~\cite{iqbal2022msanet}, HDMNet~\cite{peng2023hierarchical}, and DCAMA~\cite{shi2022dense} calculate mutual query and support correlation in multi-scale to produce multi-level correlation score map. These maps are fused to provide with a strong segmentation guidance.
However, all prior works on FSS works evaluate only cases when the number of supports ranges from 1 to 5.

\section{Preliminary Studies}
\label{sec:preliminary studies}

%
%
%
%
Intuitively, for $N$-shot FSS, readers may believe that adding more support images (i.e., increasing $N$) can lead to consistent segmentation improvements. 
However, we experimentally find that a bold increase of $N$ does not guarantee a performance gain.
We tested the SOTA FSS approach DCAMA~\cite{shi2022dense} on COCO-20$^i$~\cite{nguyen2019feature} fold 1-4. 
DCAMA is a typical correlation-based FSS framework,
%
for generality, we tested DCAMA with three common backbones, ResNet-50~\cite{hariharan2014simultaneous}, ResNet-101~\cite{hariharan2014simultaneous}, and Swin-Transforme-Base~\cite{liu2021swin}. 
%
%
We plot the mask mIoU against the number of shots in Fig.~\ref{fig: preliminary_exp_result} (left). 
Initially, the performance arises as expected, but from $N=5$, the performance improvements narrow down and even may turn negative.

Let us think about the reason. 
Although the support image(s) and the query image are in the same category, intra-class instances can have significant visual differences. 
\begin{figure}
    \centering   \includegraphics[width=0.48\textwidth]{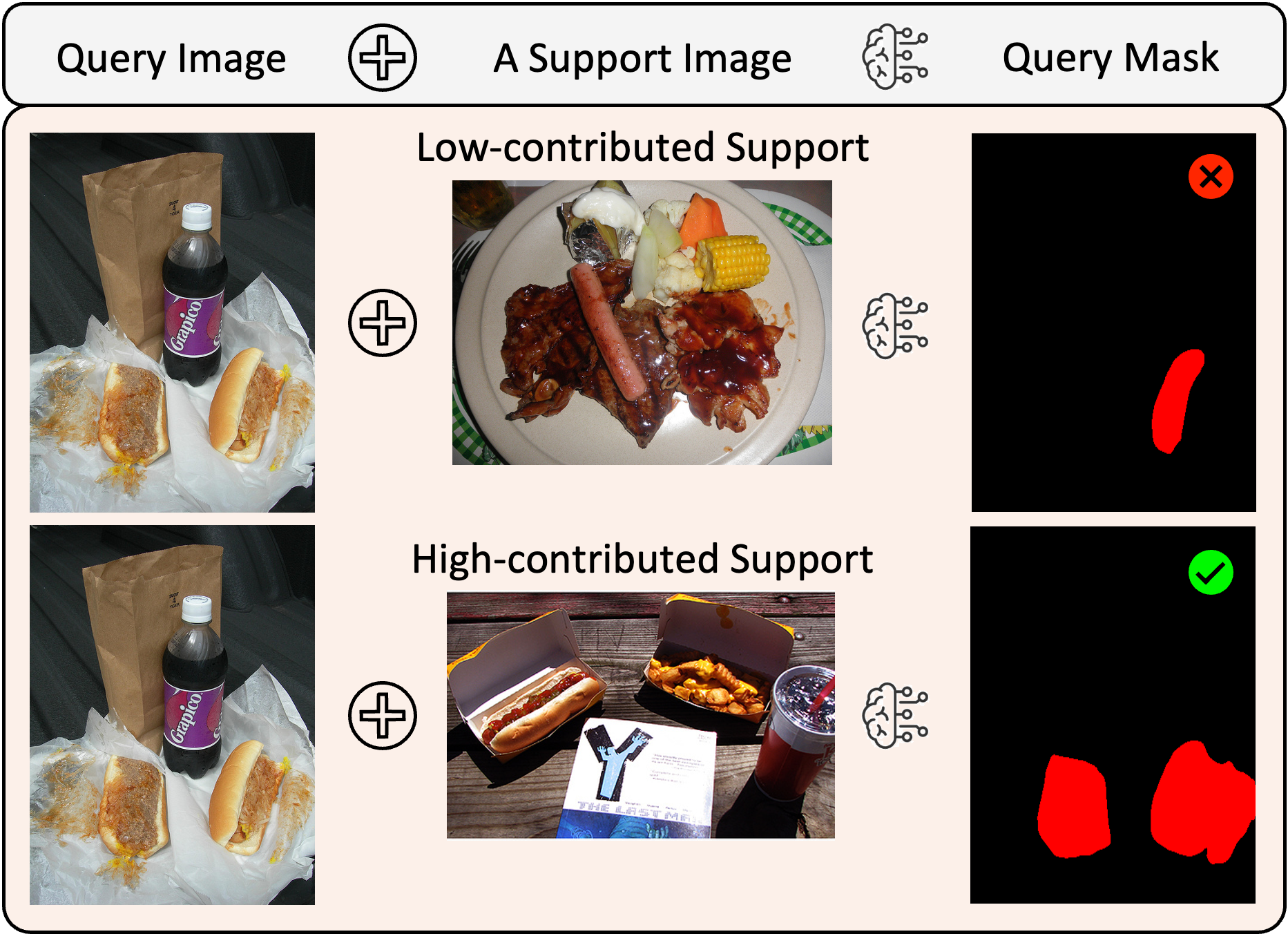}
    \caption{The supports in the same category can have significant visual differences, different contributions to the query lead to different mask results.}
    \label{fig: hotdog}
\end{figure}
For example, see Fig.~\ref{fig: hotdog}, both supports lie in the `hotdog' category, but the sausage folded in the bread and packed in a box (bottom) contributes high to the query, while the sausage placed in the brunch (top) contributes low.
unfortunately, our support set is always constructed by randomly selected images, without data filtering or refinement. 
Even if we use some data pre-processing techniques (e.g., choose supports by image visual similarities~\cite{lowe2004distinctive, dalal2005histograms, avrithis2014hough} or embedding distances~\cite{simonyan2014very, oquab2023dinov2, radford2021learning}), as $N$ gets larger, the support set inevitably gets noisy and chaotic.
Given more supports, the useful visual information increases, but so does the useless information. 
%
The big and noisy support pool challenges existing FSS methods, see more results of HDMNet~\cite{min2021hypercorrelation}(CVPR'23), SCCAN~\cite{xu2023self}(ICCV'23) and MSANet~\cite{iqbal2022msanet}(Arxiv'22) in Tab.~\ref{table:dilution} (top). These networks cannot exclusively focus on the high-contributed supports and can be negatively influenced by the low-contributed ones as $N$ increases.
We call this problem \textit{\textbf{support dilution}}. 
%
\begin{table}[t]
\centering
\caption{mIoU(\%) results of three FSS methods. * stands for experiments with the upper-bound support.}
\label{table:dilution}
\resizebox{0.45\textwidth}{!}{
\begin{tabular}{c||cccccc}
\hline
\multirow{2}{*}{Methods} & \multicolumn{5}{c}{shots}
        \\ \cline{2-7}
        & \multicolumn{1}{c|}{1}     & \multicolumn{1}{c|}{5}     & \multicolumn{1}{c|}{10}    & \multicolumn{1}{c|}{30}   & \multicolumn{1}{c|}{50} & 70    \\ \hline
HDMNet  & \multicolumn{1}{c|}{40.7} & \multicolumn{1}{c|}{46.2} & \multicolumn{1}{c|}{\textbf{47.3}} & \multicolumn{1}{c|}{46.7} & \multicolumn{1}{c|}{45.6} & 45.2     \\ \hline
SCCAN   & \multicolumn{1}{c|}{38.8} & \multicolumn{1}{c|}{41.8} & \multicolumn{1}{c|}{\textbf{43.1}} & \multicolumn{1}{c|}{41.9} & \multicolumn{1}{c|}{37.6} & 35.9 \\ \hline
MSANet     & \multicolumn{1}{c|}{38.9} & \multicolumn{1}{c|}{41.9} & \multicolumn{1}{c|}{\textbf{43.1}} & \multicolumn{1}{c|}{42.0} & \multicolumn{1}{c|}{37.6} & 41.1 \\ \hline\hline
*  & \multicolumn{1}{c|}{1}     & \multicolumn{1}{c|}{5}     & \multicolumn{1}{c|}{10}    & \multicolumn{1}{c|}{30}    & \multicolumn{1}{c|}{50} & 70   \\ \hline

HDMNet  & \multicolumn{1}{c|}{\textbf{56.3}} & \multicolumn{1}{c|}{52.3} & \multicolumn{1}{c|}{51.5} & \multicolumn{1}{c|}{51.2} & \multicolumn{1}{c|}{51.0} & 50.3     \\ \hline
SCCAN     & \multicolumn{1}{c|}{\textbf{61.2}} & \multicolumn{1}{c|}{54.1} & \multicolumn{1}{c|}{49.8} & \multicolumn{1}{c|}{46.2} & \multicolumn{1}{c|}{44.4} & 42.1 \\ \hline
MSANet   & \multicolumn{1}{c|}{\textbf{64.7}} & \multicolumn{1}{c|}{63.4} & \multicolumn{1}{c|}{63.3} & \multicolumn{1}{c|}{63.2} &\multicolumn{1}{c|}{63.0} & 59.7  \\ \hline
\end{tabular}}
\end{table}

Furthermore, we conduct an interesting experiment on DCAMA ResNet-50.
When inference, we use the query image itself as the support (i.e., the upper-bound support). 
When the paired images are identical, we will get the best FSS result, and this 1-shot test can provide us with a performance upper bound for reference.
We progressively increase $N$ from 1 to 70, i.e., we mix the upper-bound support with more chaotic information. 
If the network can capture and only capture the important part, it should consistently focus on the upper-bound support, and keep the performance at the highest point.
Unfortunately, see Fig.~\ref{fig: preliminary_exp_result} (right), DCAMA lacks such robustness when the information volume grows, the precision drop indicates that DCAMA fails in protecting the upper-bound support from dilution.
We find the same phenomenon on quite a lot FSS methods, quantitatively shown in Tab.~\ref{table:dilution} (bottom, w/ *).
None of these approaches can maintain the strength of the upper-bound support. The dilution problem is quite severe but has long been ignored.

\section{Methods}
\label{sec: method}
\subsection{Overview}
Based on Sec.~\ref{sec:preliminary studies}, there is a need to overcome support dilution. 
Given variable numbers of shots, we encourage the network to concentrate on high-contributed supports and avoid the negative disturbance from low-contributed supports. 

In Sec.~\ref{sec:method problem setting}, we give the problem definition.
Then, we sequentially propose three technical contributions.
In Sec.~\ref{method CF}, we design a contribution index that quantitatively estimates the amount a support truly contributes to the query. 
%
%
%
In Sec.~\ref{method SC}, we propose Symmetric Correlation (SC), which is used to protect and stabilize the contribution values of high-contributed supports against the distraction of noise, so that the query can absorb essential information and achieve better segmentation results.
More than SC, in Sec.~\ref{method SIP}, we propose an operation called Support Image Pruning, where we present a contribution-guided greedy algorithm to retrieve a small and high-quality subset. 
%
%
With the purified set, SC can better leverage its concentration ability on high-contributed supports consuming less computation costs.
Finally, in Sec.~\ref{method pipeline}, we present a detailed introduction of our pipeline, equipped with the above technical designs.
%
%
%

%

%
\subsection{Problem Setting}
\label{sec:method problem setting}
Given a query image $I_q$ containing novel objects, we feed $N$ support data $\mathcal{S} = \{(I_{S_i}, M_{S_i})\}_{i=1}^N$ into the network to provide the knowledge of the novel category, where $I_{S_i}$ represents a support image and $M_{S_i}$ represents its corresponding binary mask. Like FSS, we want to harness the information from $\mathcal{S}$ to facilitate segmenting $I_q$. 
Beyond FSS, $N$ is a random and dynamic number, and it can be quite large, depending on real-world conditions.
Given a large $S$, our goal is to construct a robust segmentation framework that can preserve the high-contributed supports and suppress the negative influence of those low-contributed supports.
\subsection{Contribution Index}
\label{method CF}
%
Support Dilution can be vividly described as, the high-contributed (i.e., visually relevant) supports actually contribute low to the query, whereas the low-contributed (i.e., relatively irrelevant) supports turn out to contribute high. 
The prerequisite task is to quantitatively estimate the real contribution of each support image $I_{S_i}$.
%
We design a contribution index. For each support, its contribution value is approximately proportional to the support-query correlation weight. 
%
%
%

%
Let us start from the standard cross attention mechanism. Suppose we have an one-to-one support-query image pair $\{I_s, I_q\}$, generally, we use a shared pretrained backbone $Enc$ to extract the support feature $x_s=Enc(I_s)$ and the query feature $x_q=Enc(I_q)$.
Then, two different feed-forward networks $f_K$ and $f_Q$ are applied to generate Key and Query: $K=f_K(x_s), Q=f_Q(x_q)$.
Following the standard scaled dot-product attention $A(K,Q)=\text{softmax}(\frac{QK^{T}}{\sqrt{d}})$  ($d$ is the feature dimension),
we can rewrite the attention matrix as a function determined by two variables $x_s$ and $x_q$:
\begin{equation}
\label{Asq1}
    A(x_s,x_q)=\text{softmax}(\frac{f_Q(x_q)f_K(x_s)^T}{\sqrt{d}}).
\end{equation}
If $I_s$ contains lots of meaningful visual knowledge, ideally, $x_s$ will contribute more in the representation space and result in a larger attention weight. 
We hence define our contribution index $\delta(\cdot)$ as:
\begin{equation}
\label{Deltas1}
    \delta(x_s)=\frac{1}{|x_s|}\sum_{i=1}^{|x_s|} \max_{j=1,...,|x_q|}A(x_s, x_q)_{[i,j]},
\end{equation}
where $|\cdot|$ obtains the token amount, and the subscript $[i,j]$ is used to access $i_{\text{th}}$-row-$j_{\text{th}}$-column entry of the matrix.
In Eq.~\ref{Deltas1}, the importance of each token in $x_s$ is the maximum attention weight it has among all tokens of $x_q$, and $\delta(x_s)$ indicates the overall mean value across $x_s$'s tokens.
With Eq.~\ref{Deltas1}, when a new support comes, we can calculate its contribution value, and if there are two supports, we can fairly compare their contributions by leveraging Eq.~\ref{Deltas1} twice.

We then extend the one-query-one-support case to the one-query-multi-support case. 
Given a big support set $S$ with $N$ supports $\{I_{S_i}\}_{i=1}^{N}$, similarly, we use $Enc$ to obtain support features $X_S=[x_{S_{1}},...,x_{S_N}]$, where $X_S$ is the concatenation of $N$ support features. 
To calculate the attention value for each $x_{S_i}$, we extend Eq.~\ref{Asq1} to:
\begin{equation}
\label{Asq2}
\begin{aligned}
    A_i(X_S,x_q)=\left[\text{softmax}(\frac{f_q(x_q)f_k(X_{S})^T}{\sqrt{d}}))\right]_{(:,[\text{head}_i:\text{tail}_i])}, &\\
    i=1,...,N&.
\end{aligned}
\end{equation}
where the subscript ${(:,[\text{head}:\text{tail}])}$ denotes the slice (i.e., submatrix) of extracting the columns from $\text{head}$ to $\text{tail}$, $\text{head}_i=\sum_{j=1}^{i-1} |x_{S_j}|$, $\text{tail}_i=\text{head}_i+|x_{S_i}|$. 
Note that among $X_S$, some features are high-contributed while most others are low-contributed, but the high-contributed ones may turn out to contribute less due to the dilution problem. 
To see how much each $x_{S_i}$ actually contributes to $x_q$, we extend Eq.~\ref{Deltas1} to a multi-support contribution index:
\begin{equation}
\begin{aligned}
    \label{Deltas2}
    \delta(x_{S_i})=\frac{1}{|x_{S_i}|}\sum_{i=1}^{|x_{S_i}|} \max_{j=1,...,|x_q|}A_i(X_S, x_q)_{[i,j]}, &\\
    i=1,...,N&.
\end{aligned}
\end{equation}
Eq.~\ref{Deltas2} can help us to
find if a high-contributed support dilutes and how bad it dilutes. For example, if we have a high-contributed support and a low-contributed support, the former one should result in a higher contribution value with Eq.~\ref{Deltas2}, if not, the relative deviation can reveal the degree of dilution.

\subsection{Symmetric Correlation}
\begin{figure}%
    \centering
    \includegraphics[width=0.46\textwidth]{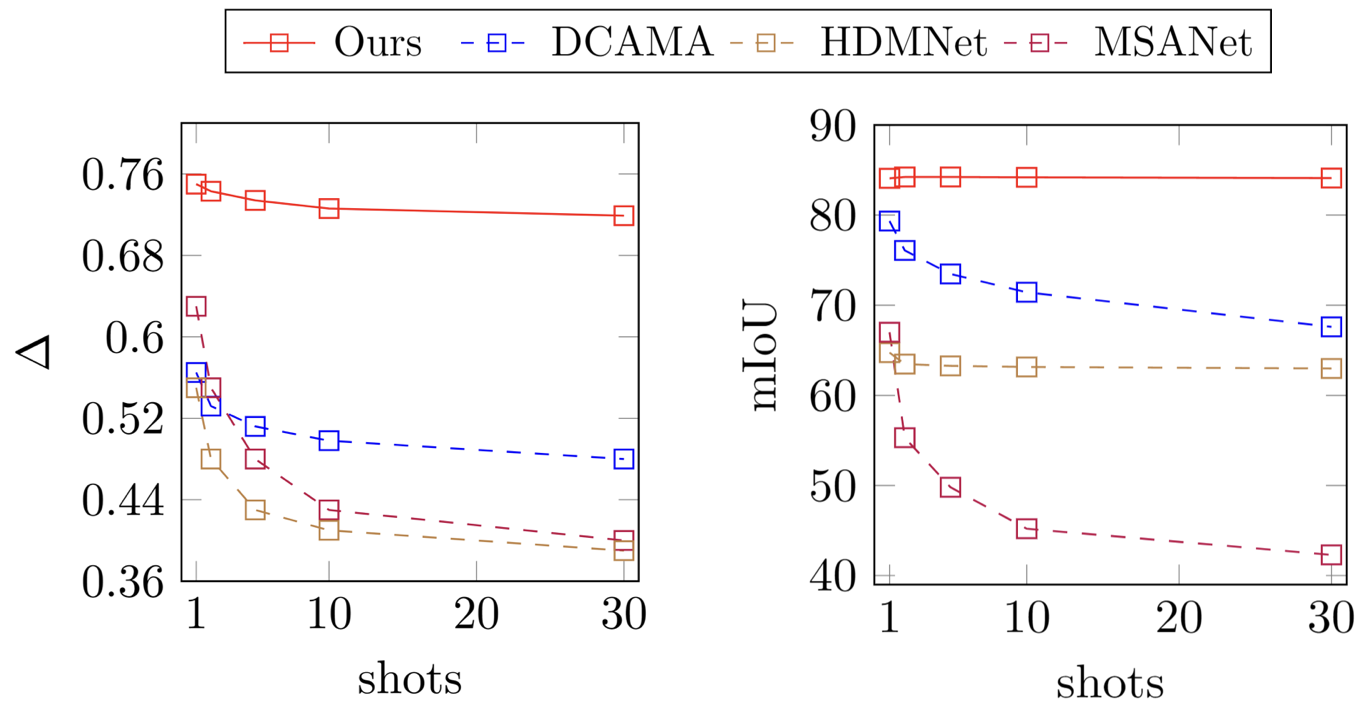}
    \caption{Left: The deviation value $\Delta$ v.s. the number of shots $N$. Right: segmentation mIoU v.s. the number of shots $N$. Experiments on COCO-20$^i$ Fold 1 under the upper-bound setting.}%
    \label{fig:deviation}%
\end{figure}
\label{method SC}
To make full use of the support(s) for segmenting the query, existing FSS methods~\cite{shi2022dense, iqbal2022msanet, xu2023self,min2021hypercorrelation, hu2019attention} study various strategies to extract support-query correlations.
In this section, we study why and how those correlation mechanisms lead to the support dilution problem. 
Then, we propose a simple solution, Symmetric Correlation (SC), which can successfully alleviate support dilution.

%
%
Ideally, the high-contributed support features should dominate the correlation scores (e.g., attention values), and the low-contributed support features should result in low correlation scores.
In other words, we expect a high-contributed $x_{S_i}$ to get a relatively large contribution value in Eq.~\ref{Deltas2}, depending on the visual similarity between $I_{S_i}$ and $I_q$.
%
%
Undoubtedly, the upper-bound support (i.e., use the query itself as the support) will result in the highest contribution index.
Suppose it generates feature $x_{S_1}$ and other noisy supports generate features $x_{S_2},...,x_{S_N}$, a capable correlation module should always keep $\delta(x_{S_1})$ at the upper-bound value no matter how large $N$ gets. 
Specifically, we use $\bar{\delta} = \frac{1}{N-1} \sum_{i=2}^N \delta(x_{S_i})$ to approximate the less-important contribution index, and use the deviation $\Delta=\delta(x_{S_1}) - \bar{\delta}$ to measure the degree that the upper-bound support feature standing out from other features. 
For $\Delta$, we certainly prefer a consistently large number, which means that the highest-contributed feature $x_{S_1}$ can effectively suppress other support features and dominate the support-query correlation, demonstrating the robustness of the correlation module against the distraction of noisy information.

Let us take a quick glimpse of how previous FSS methods work. Experiments are conducted on COCO-20$^i$ Fold 1. 
Illustrated in Fig.~\ref{fig:deviation} (left) dashed lines, the $\Delta$s are initialized at small values, and as $N$ increases from 1 to 30, the lines drastically go down. Obviously, none of these FSS methods can stop $x_{S_1}$ from diluting in the noisy feature pool, the high-contributed support no longer keeps its power in the correlation learning. 
Importantly, the dilution level determines the mask precision. As illustrated in Fig.~\ref{fig:deviation} (right) dashed lines, for every method, accompanied by the decrease of $\Delta$, there is a simultaneous segmentation performance drop since $x_{S_1}$ is severely weakened.

%
%
To alleviate the dilution problem, we propose a simple but effective correlation module, called Symmetric Correlation (SC). 
We design SC with a constraint that, the attention weight reaches the maximum \textbf{\textit{when and only when}} the support feature and the query feature are identical, i.e., when we input the upper-bound support. 
With this upper-bound constraint, we modify the original one-query-one-support correlation (Eq.~\ref{Asq1}) to:
\begin{equation}
\begin{aligned}
\label{SymAtt1}
    A(x_s, x_q)=\text{softmax}(\frac{f(x_s)f(x_q)^T}{\sqrt d}),&\\
    f(x) = \frac{f_1(x)f_2(x)}{\|f_2(x)\|_2}.&
\end{aligned}
\end{equation}
Similarly, we reformulate the one-query-variable number-support correlation (Eq.~\ref{Asq2}) as:
\begin{equation}
\begin{aligned}
\label{SymAtt2}
     A_i(X_S,x_q)=\left[\text{softmax}(\frac{f(x_q)f(X_S)^T}{d}))\right]_{(:,[\text{head}_i:\text{tail}_i])}, &\\
    f(x) = \frac{f_1(x)f_2(x)}{\|f_2(x)\|_2},i=1,...,N, &
\end{aligned}
\end{equation}
where $\text{head}_i$ and $\text{tail}_i$ serve the same as in Eq.~\ref{Asq2}.

In SC (Eq.~\ref{SymAtt2}), we make two improvements. 
Firstly, to meet the upper-bound constraint, we use the same network $f$ to generate both the Key and the Query, symmetry guarantees the attention function to have an unique maximum point.
%
Secondly, we normalize the Key and the Query before matrix multiplication. 
The normalization operation is composed of two parts, the magnitude part $f_1(x)$ and the angle part $\frac{f_2(x)}{\|f_2(x)\|_2}$. 
The magnitude part predicts the absolute importance of the Key and the Query (i.e., foreground/background, or we can say objectness), while the angle part predicts the relative importance between the Key and the Query (i.e., support-query relation, or we can say similarity). 
In this way, SC learns intra-correlation and inter-correlation simultaneously.
Moreover, in the one-query-one-support case (Eq.~\ref{SymAtt1}), $A(x_s,x_q)=A(x_q,x_s)$, this can guarantee the shuffling-invariant stability of SC against disturbance~\cite{zhang2021few}.

We illustrate how we preserve and strengthen the high-contributed support feature in Fig.\ref{fig:deviation} (left) solid line. Compared to SOTA FSS methods, our deviation value $\Delta$ starts from a higher point and is less impaired when $N$ gets larger, which means $x_{S_1}$ consistently gets much more attention weight than those less-important supports.
Since SC can enhance the strength of high-contributed supports over low-contributed supports, we achieve better segmentation performance and retain the mIoU value at a higher bound, as shown in Fig.~\ref{fig:deviation} (right) solid line.
%
%
\subsection{Support Image Pruning}
\label{method SIP}
In Sec.\ref{method SC}, we propose the Symmetric Correlation (SC) to alleviate the support dilution problem.
Take a step forward, in real-world applications, $N$ can be quite large. This brings two extra challenges. First, without manual selection, the numbers of high-contributed and low-contributed supports can be very imbalanced, the overwhelming low-contributed information can sum up to a considerable volume of noise then harm SC. Second, the computation cost of SC is determined by $N$, when $N$ is large and most of the supports are low-contributed, we will pay heavy and redundant computation effort for the big support set.  

Toward the two challenges, we propose an operation called Support Image Pruning. Pruning means deleting the useless items. 
Through the pruning operation, we eliminate the valueless supports before calculating attention, so SC will concentrate only on the relevant and meaningful supports with less computation costs.

Given the original support set $S=\{I_{S_i}\}_{i=1}^N$ (we omit the support mask for briefness) and the extracted support features $X_S$, our goal is to retrieve a subset $S'$ consisting of $N'$ ($N'<N$) support images with feature $X_{S'}$, such that their overall contribution $\sum_{x_{S'} \in X_{S'}} \delta(x_{S'})$ can be maximized, then the rest $N-N'$ useless supports will be discarded. 
This is a subset retrieval problem, and the retrieval principle can be mathematically formulated as:
\begin{equation}
\begin{aligned}
\label{Pruning}
    S' &= \mathop{\arg \max}_{{S'}\subset{S}}
    \sum_{x_{S'} \in X_{S'}} \delta(x_{S'}) =  \mathop{\arg \max}_{{S'}\subset{S}} \sum_{i=1}^{N'} \delta(x_{{S'}_i}) \\
    & =\mathop{\arg \max}_{{S'}\subset{S}} \sum_{i=1}^{N'}
    \frac{1}{|x_{{S'}_i}|}\sum_{j=1}^{|x_{{S'}_i}|} \max_{k=1,...,|x_q|} A_i(X_{S'}, x_q)_{[j,k]},
\end{aligned}
\end{equation}
which is based on our contribution index $\delta(\cdot)$ in Eq.~\ref{Deltas2}.

Finding $S'$ is not easy, there are two computation difficulties as $N$ grows. 
Firstly, obtaining $\sum_{x_{S'} \subset X_{S'}} \delta(x_{S'})$ requires attention computation for all the features in $X_{S'}$. It has the same amount of calculation effort as we consume in SC, thus does not gain any efficiency improvement. 
Secondly, exhaustively enumerating all possible $S' \subset S$ requires $O(N!)$ time complexity. As $N$ gets larger, the huge traverse costs will make this subset retrieval task too heavy to be solved.

For the two difficulties, we propose two solutions, respectively. 
For the first problem, by Jensen's inequality, given Eq.~\ref{Pruning}, we have
\begin{equation}
\begin{aligned}
\label{Pruning1st}
    \frac{1}{|x_{{S'}_i}|}\sum_{j=1}^{|x_{{S'}_i}|} \max_{k=1,...,|x_q|} A_i(X_{S'}, x_q)_{[j,k]} \geq&\\
     f(\frac{1}{|x_{{S'}_i}|}\sum_{j=1}^{|x_{{S'}_i}|}x_{{{S'}_i}_{[j]}})f(\frac{1}{|x_q|}\sum_{j=1}^{|x_q|}x_{q_{[j]}}),&
\end{aligned}
\end{equation}
where $f$ is the SC normalization function in Eq.~\ref{SymAtt2}, and $\cdot_{[i]}$ or $\cdot_{[j]}$ is the operation to get the $i_{\text{th}}$ or $j_{\text{th}}$ token.
We can use the lower bound in Eq.~\ref{Pruning1st} to replace the heavy SC computation in Eq.~\ref{Pruning}, 
then for each support image, instead of calculating the correlation for every $x_{{{S'}_i}_{[j]}}$, now we only need to calculate for the average token and apply $f$ only once. 
We can then transform the retrieval principle in Eq.~\ref{Pruning} to:
\begin{equation}
\begin{aligned}
\label{Pruning2st}
    S' &= \mathop{\arg \max}_{{S'}\subset{S}} 
    \underbrace{
    \sum_{i=1}^{N'}
    f(\frac{1}{|x_{{S'}_i}|}\sum_{j=1}^{|x_{{S'}_i}|}x_{{{S'}_i}_{[j]}})f(\frac{1}{|x_q|}\sum_{j=1}^{|x_q|}x_{q_{[j]}})}_{\theta(X_{S'})}.
\end{aligned}
\end{equation}
%
%
Dealing the second difficulty, we design a greedy algorithm.
Instead of traversing a big subset from $S$, we gradually push one then one item to the subset $S'$ and finally fill it with $N'$ items. 
Specifically, in each iteration, we retrieve one support image that can satisfy Eq.~\ref{Pruning2st} the most, so the contribution sum can be approximately maximized with a much lower time complexity $O(N' \times N)$. Please see Algo.~\ref{algo} for our full Support Image Pruning operation, where we retrieve $S'$ and the corresponding image index list $index$.
\begin{algorithm}[H]
\DontPrintSemicolon
  \KwInput{A set $S$ including $N$ support images $I_{S_1},...I_{S_N}$, a query image $I_q$, a feature encoder $Enc$, a function $f$, and a parameter  $N'$, $N' < N$.}
  \KwOutput{A subset $S' \subset S$ with $N'$ support images.}
  $i\leftarrow 0; S' \leftarrow \emptyset; X_S=Enc(S);x_q=Enc(I_q)$
  
  $index \leftarrow \emptyset$
  
   \For{$i < N'$}
    {
        $I_s\leftarrow$ $null$; $sum = -inf$; $n=1$; $n'=n$
        
        \For{$n \leq N$ and $n \notin index$}    
        { 
            $X \leftarrow X_S[index \cup \{n\}]$
            
            ${sum}' \leftarrow 
            \theta(X)$
            
            \If{${sum}'>sum$}
            {
                $n'=n$;

                $sum\leftarrow {sum}'$
            }
        }
        $S' \leftarrow S' \cup \{I_{S_{n'}}\}$; $index \leftarrow index \cup n'$

        $i \leftarrow i + 1$
    }
\caption{Algorithm for retrieving subset $\mathcal{S}'$}
\label{algo}
\end{algorithm}
\begin{figure*}[ht]
    \includegraphics[width=\linewidth]{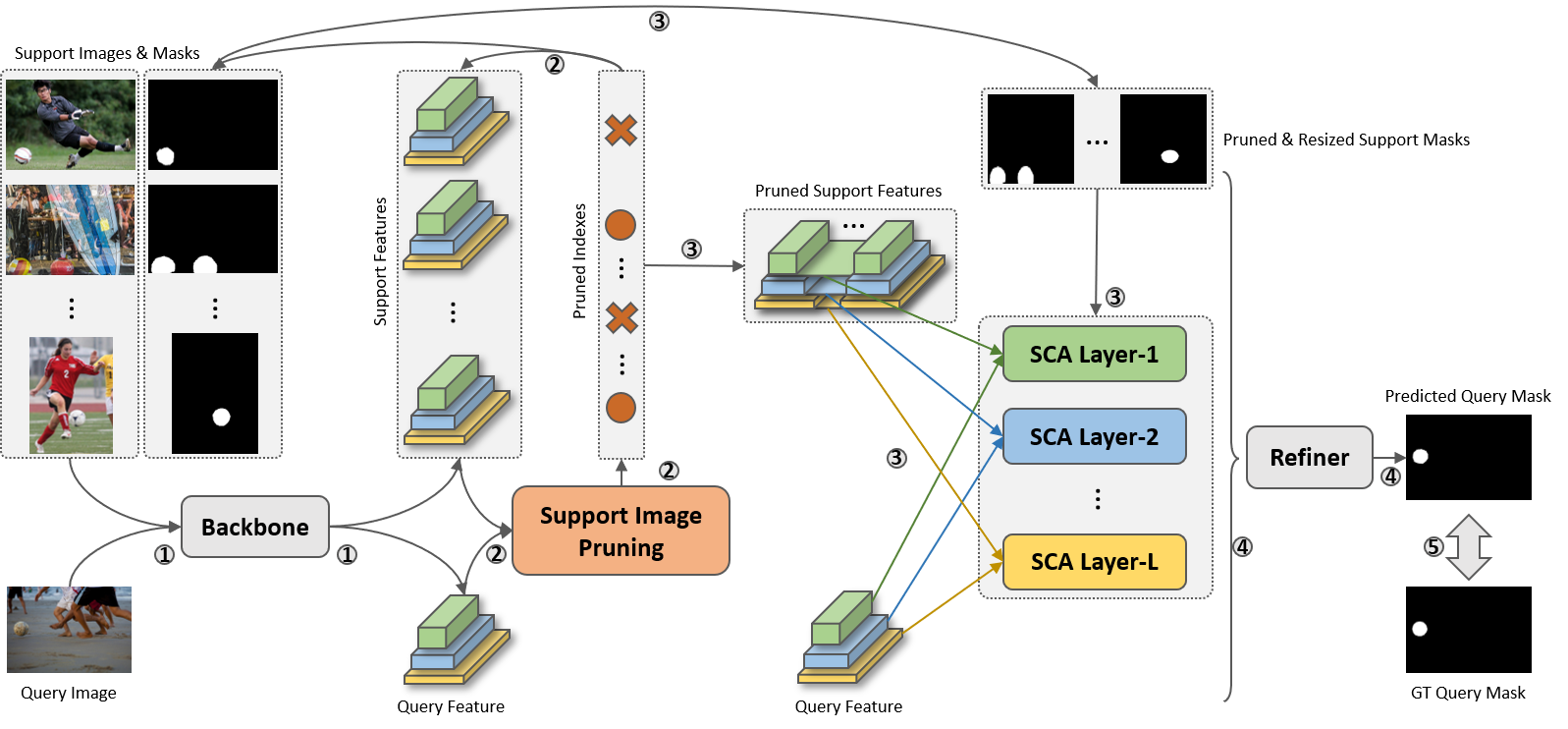}
    \caption{The pipeline of our network. We introduce the flow \ding{192}-\ding{196} in Sec.~\ref{method pipeline}. The most critical parts, \ding{193} and \ding{194}, are developed against support dilution. \ding{193}: Support Image Pruning (Sec.~\ref{method SIP}) is deployed to select high-contributed supports (indexed by the $\circ$ icon) and abandon low-contributed supports (indexed by the $\times$ icon). \ding{194}: A novel correlation module, Symmetric Correlation (Sec.~\ref{method SC}), is multi-layer applied to preserve and enhance the high-contributed features.}
\label{fig: pipeline}
\end{figure*}
\subsection{Pipeline}
\label{method pipeline}
In this section, we introduce the detailed pipeline of our network, which is illustrated in Fig.~\ref{fig: pipeline}. We show the flow step by step (\ding{192}-\ding{197}).

\ding{192}: Given the support set $S = \{(I_{S_i}, M_{S_i})\}_{i=1}^N$, a shared backbone $Enc$ (e.g., ResNet, Swin-Transformer, etc.) is applied on the support images $\{I_{S_i}\}_{i=1}^N$ and the query image $I_q$ to extract multi-layer high-dimension features:
\begin{equation}
\begin{aligned}
    \{X_{S}^l\}_{l=1}^L = [\{x_{S_1}^l\},...,\{x_{S_N}^l\}]_{l=1}^L &\leftarrow Enc(\{I_{S_i}\}_{i=1}^N), \\
    \{x_q^l\}_{l=1}^{L} &\leftarrow Enc(I_q),
\end{aligned}
\end{equation}
where $L$ is the number of feature layers (in Fig.~\ref{fig: pipeline}, $L=3$, we color the three layers in green, blue and yellow).

\ding{193}: We adopt the Support Image Pruning (Sec.~\ref{method SIP}) to obtain a small subset $S'$ from $S$. 
Specifically, in this step, the multi-layer retrieval principle is
\begin{equation}
\begin{aligned}
        S' &= \mathop{\arg \max}_{{S'}\subset{S}} 
    \frac{1}{L}\sum_{l=1}^{L} \theta(X_{S'}^l), |S'|=N'.
\end{aligned}
\end{equation}
We average the contributions of each layer and use the mean value as our optimizing target.
Following Algo.~\ref{algo}, we discard some low-contributed supports and only keep $N'$ informative ones, the pruned indexes (marked with $\circ$ or $\times$ icons) can be used to filter the support features and support masks.

\begin{figure}[ht]
    \includegraphics[width=0.98\linewidth]{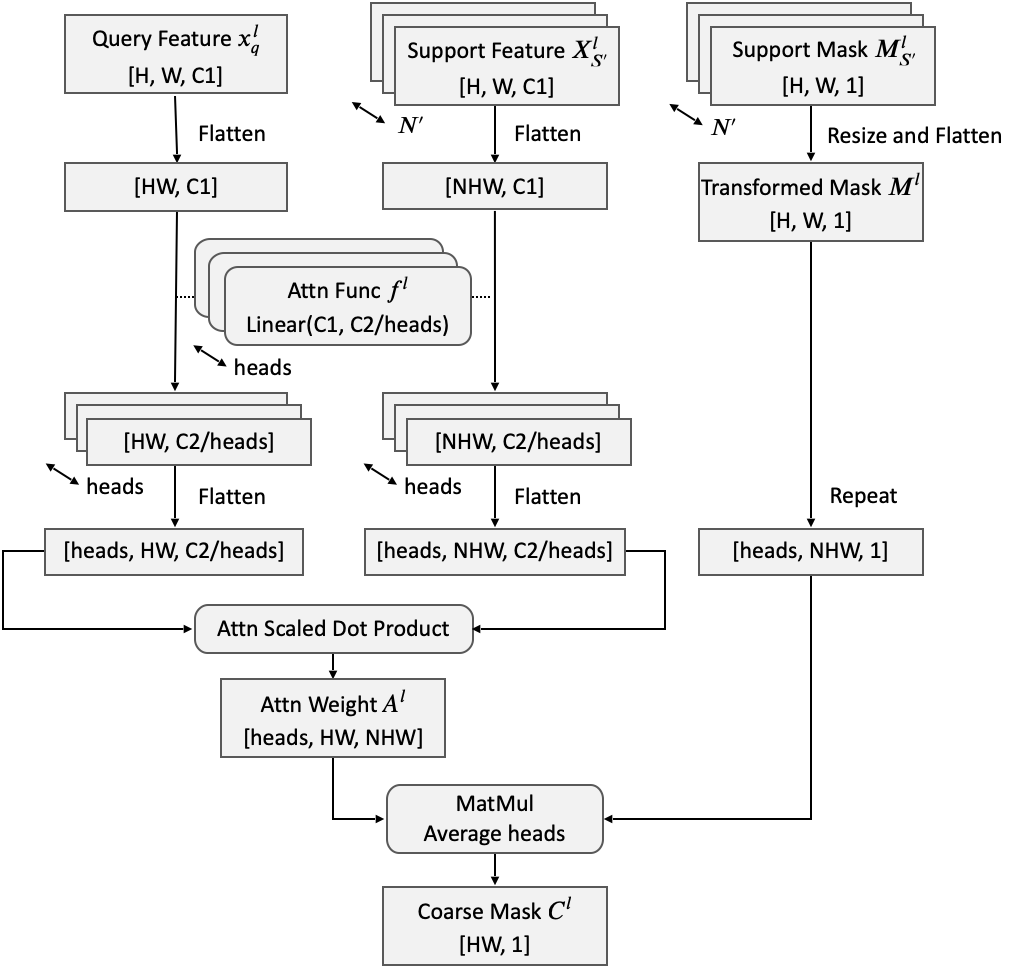}
    \caption{The dataflow and intermediate feature shapes of the $l_{\text{th}}$-layer SC (shown as `SCA Layer-$l$' in Fig.~\ref{fig: pipeline}).}
\label{fig: attn}
\end{figure}

\ding{194}: Now we obtain the concatenated pruned features and the resized pruned masks. 
We then build multi-layer SC modules (Sec.~\ref{method SC}).
Following Eq.~\ref{Deltas2}, the SC calculation for the $l_{\text{th}}$ feature of the $i_{\text{th}}$ support can be formulated as:
\begin{equation}
\begin{aligned}
\label{SymAtt3}
 A_i^l(X_{S'}^l,x_q^l)=\left[\text{softmax}(\frac{f^l(x_q^l)f^l(X_{S'}^l)^T}{d}))\right]_{(:,[\text{head}_i:\text{tail}_i])}, &\\
    f^l(x) = \frac{f_1^l(x)f_2^l(x)}{\|f_2^l(x)\|_2},i=1,...,N', &
\end{aligned}
\end{equation}
where $\text{head}_i$ and $\text{tail}_i$ serve the same as in Eq.~\ref{Asq2}. Please see Fig.~\ref{fig: attn} for more details of the $l_{\text{th}}$-layer multi-head SC.
By applying SCs, we obtain multi-layer correlation weights $\{A^l\}_{l=1}^{L}$, and each support-specific correlation weight (Eq.~\ref{SymAtt3}) can be extracted by slicing: 
\begin{equation}
    A_i^l(X_{S'}^l,x_q^l) = {A^l}_{(:,[\text{head}_i:\text{tail}_i])}.
\end{equation}
$\{A^l\}_{l=1}^{L}$ can preserve the information of high-contributed supports from dilution, and suppress the noise of the low-contributed supports.

\ding{195}: With $\{A^l\}_{l=1}^{L}$, we apply a Refiner to obtain the segmentation result.
%
For the $l_{\text{th}}$ layer, the corresponding coarse mask is generated by $C^l = A^l \cdot M^l$, shown in Fig.~\ref{fig: attn}.
%
%
Our Refiner harnesses the top-down fusion to aggregate coarse masks of neighbour layers.
In each top-down step, we apply bilinear interpolation $U^{l-1} = \text{Upsample}(C^l)$ to align $C^l$ with the size of $C^{l-1}$, then we refine $C^{l-1}$ by a 2D convolution $F^{l-1} = \text{conv}^{l-1}(\text{concat}[U^{l-1},C^{l-1}])$.
%
%
We repeat the top-down process to fuse two consecutive layers' coarse predictions until we obtain the second-last-layer $F^{2}$. 
In the last step, we obtain the final binary output $F^1$ by $F^1=\text{conv}^1(\text{concat}[\text{Upsample}(F^2), x_q^1, \text{AvgPool}(X_{S'}^1)])$.

\ding{196}: Following common FSS methods, our pipeline is supervised by the Cross Entropy loss\cite{good1952rational}. Given the ground-truth label $\hat{F^1}$ of the query image, the loss function is:
\begin{equation}
    \mathcal{L}=-\frac{1}{|F^1|}\sum_{x\in F^1\\\hat{x}\in \hat{F^1}}((x)log(l(\hat{x})) + (1-x)log(1-\hat{x})).
\end{equation}

In this section, we present a simple and easy-to-understand pipeline.
In fact, our proposed techniques, i.e., Symmetric Correlation (Sec.~\ref{method SC}) and Support Image Pruning (Sec.~\ref{method SIP}), can be plugged into many FSS networks to deal with support dilution, showing their generality and practicality. Plug-and-play experiment results can be found in Sec.~\ref{exp: plug-and-play exp}. 

%
%

\section{Experiment Results}
\label{exp}
\subsection{Overview}
We conduct extensive experiments to validate the effectiveness of our approach to deal with support dilution in FSS.

In Sec.~\ref{exp: imple}, we introduce our implementation details.

In Sec.~\ref{exp: normal exp}, for fair comparisons with SOTA FSS methods, we show benchmark results on COCO-20$^{i}$~\cite{nguyen2019feature} and PASCAL-5$^i$~\cite{shaban2017one}, the data statistics can be found in 
Tab.~\ref{tab: statistics}. 
We gradually increase $N$ from 1 to 70, and present both quantitative and qualitative results. 
For quantitative evaluation, we report the mean Intersection-over-Union (mIoU), which is calculated as $\frac{\sum_{i=0}^{c} \text{mIoU}_i}{c}$, where $c$ is the number of classes in the test fold and $\text{mIoU}_i$ represents the mIoU value of the ${i}^{th}$ class.
\begin{figure*}
    \centering
    \hspace*{-0.5cm}
        \begin{tikzpicture}
        \tikzstyle{every node}=[font=\small]
            \begin{groupplot}[
                group style={
                    group name=my plots,
                    group size=6 by 4,
                    ylabels at=edge left
                },
                legend pos=south east,
                footnotesize,
                width=\textwidth / 5,
                height=\textwidth / 5,
                tickpos=left,
                ytick align=outside,
                xtick align=outside,
                enlarge x limits=false ,
            ]
            \nextgroupplot[title={Res-50, Fold1, },ylabel={mIoU, COCO}]
            \addplot[color=blue,mark=square, mark size=1.3pt]
                    coordinates{
                     (1,41.17)(5,44.67)(10,44.97)(30, 44.32)(50, 42.93)(70, 41.91)
                    };\label{plots:DCAMA}
            \addplot[color=orange,mark=square,mark size=1.3pt]
                    coordinates{
                     (1,41.49)(5,50.58)(10,53.11)(30, 55.40)(50, 55.42)(70, 56.11)
                    };\label{plots:Ours}
            \addplot[color=purple,mark=square,mark size=1.3pt]
                    coordinates{
                     (1, 38.9)(5,41.9)(10,43.1)(30, 42.0)(50, 37.6)(70, 41.1)
                    };\label{plots:MSANet}
            \coordinate (top) at (rel axis cs:0,1);
           \nextgroupplot[title={Res-101, Fold1}]
            \addplot[color=blue,mark=square,mark size=1.3pt]
                    coordinates{
                     (1,42.39)(5,48.30)(10,48.30)(30,48.16)(50, 47.67)(70, 46.77)
                    };\label{plots:DCAMA}
            \addplot[color=orange,mark=square,mark size=1.3pt]
                    coordinates{
                     (1,40.79)(5,51.77)(10,53.58)(30, 55.88)(50, 56.56)(70, 56.51)
                    };\label{plots:Ours}
            \addplot[color=purple,mark=square,mark size=1.3pt]
                    coordinates{
                     (1, 40)(5, 41.5)(10, 41.6)(30, 37.1)(50, 42.4)(70, 36.4)
                    };\label{plots:MSANet}
                    
            \nextgroupplot[title={Swin, Fold1}]
            \addplot[color=blue,mark=square,mark size=1.3pt]
                    coordinates{
                     (1,47.56)(5,54.91)(10,56.22)(30,58.88)(50, 59.15)(70, 59.67)
                    };\label{plots:DCAMA}
            \addplot[color=orange,mark=square,mark size=1.3pt]
                    coordinates{
                     (1,47.64)(5,56.29)(10,58.62)(30,60.92)(50, 60.90)(70, 61.42)
                    };\label{plots:Ours}
            \nextgroupplot[title={Res-50, Fold2}]
            \addplot[color=blue,mark=square,mark size=1.3pt]
                    coordinates{
                     (1,43.62)(5,50.76)(10,51.97)(30,52.46)(50, 52.32)(70, 51.70)
                    };\label{plots:DCAMA}
            \addplot[color=orange,mark=square,mark size=1.3pt]
                    coordinates{
                     (1,43.45)(5,53.29)(10,55.54)(30,56.74)(50, 57.85)(70, 58.44)
                    };\label{plots:Ours}
             \addplot[color=purple,mark=square,mark size=1.3pt]
                    coordinates{
                     (1, 49.5)(5, 49.6)(10, 48.1)(30, 50.0)(50, 51.8)(70, 50.1)
                    };\label{plots:MSANet}
            \nextgroupplot[title={Res-101, Fold2}]
            \addplot[color=blue,mark=square,mark size=1.3pt]
                    coordinates{
                     (1,47.98)(5,55.85)(10,58.62)(30, 58.68)(50, 58.48)(70, 60.43)
                    };\label{plots:DCAMA}
            \addplot[color=orange,mark=square,mark size=1.3pt]
                    coordinates{
                     (1,46.35)(5,58.36)(10,60.85)(30, 63.84)(50, 63.95)(70, 64.15)
                    };\label{plots:Ours}
             \addplot[color=purple,mark=square,mark size=1.3pt]
                    coordinates{
                     (1, 50.6)(5, 52.5)(10, 55.5)(30, 56.4)(50, 53.2)(70, 52.9)
                    };\label{plots:MSANet}
                    
            \nextgroupplot[title={Swin, Fold2}]
             \addplot[color=blue,mark=square,mark size=1.3pt]
                    coordinates{
                     (1,53.51)(5,59.77)(10,61.84)(30,62.67)(50, 63.15)(70, 63.65)
                    };\label{plots:DCAMA}
            \addplot[color=orange,mark=square,mark size=1.3pt]
                    coordinates{
                     (1,52.20)(5,60.20)(10,61.39)(30,64.17)(50, 64.78)(70, 64.84)
                    };\label{plots:Ours}
           \nextgroupplot[title={Res-50, Fold3}, ylabel={mIoU, COCO}]
            \addplot[color=blue,mark=square,mark size=1.3pt]
                    coordinates{
                     (1,43.14)(5,49.78)(10,51.09)(30,50.75)(50, 50.28)(70, 50.75)
                    };\label{plots:DCAMA}
            \addplot[color=orange,mark=square,mark size=1.3pt]
                    coordinates{
                     (1,40.38)(5,53.89)(10,56.28)(30,56.98)(50, 57.46)(70, 57.79)
                    };\label{plots:Ours}
            \addplot[color=purple,mark=square,mark size=1.3pt]
                    coordinates{
                     (1, 41.7)(5, 45.5)(10, 47.0)(30, 43.7)(50, 44.2)(70, 47.0)
                    };\label{plots:MSANet}
             \nextgroupplot[title={Res-101, Fold3}]
             \addplot[color=blue,mark=square,mark size=1.3pt]
                    coordinates{
                     (1,45.17)(5,52.92)(10,54.67)(30,53.79)(50, 52.61)(70, 51.91)
                    };\label{plots:DCAMA}
            \addplot[color=orange,mark=square,mark size=1.3pt]
                    coordinates{
                     (1,44.97)(5,55.64)(10,57.49)(30,58.54)(50, 58.89)(70, 58.95)
                    };\label{plots:Ours}
            \addplot[color=purple,mark=square,mark size=1.3pt]
                    coordinates{
                     (1, 45.3)(5, 48.7)(10, 45.7)(30, 48.0)(50, 46.1)(70, 46.3)
                    };\label{plots:MSANet}
            \nextgroupplot[title={Swin, Fold3}]
            \addplot[color=blue,mark=square,mark size=1.3pt]
                    coordinates{
                     (1,50.81)(5,60.32)(10,61.82)(30, 62.59)(50, 62.97)(70, 60.63)
                    };\label{plots:DCAMA}
            \addplot[color=orange,mark=square,mark size=1.3pt]
                    coordinates{
                     (1,50.40)(5,61.43)(10,62.65)(30, 62.97)(50, 63.03)(70, 63.29)
                    };\label{plots:Ours}
            \nextgroupplot[title={Res-50, Fold4}, ]
             \addplot[color=blue,mark=square,mark size=1.3pt]
                    coordinates{
                     (1,40.18)(5,46.70)(10,46.75)(30,46.56)(50, 45.07)(70, 44.60)
                    };\label{plots:DCAMA}
            \addplot[color=orange,mark=square,mark size=1.3pt]
                    coordinates{
                     (1,39.20)(5,49.21)(10,51.43)(30,53.32)(50, 54.21)(70, 54.72)
                    };\label{plots:Ours}
             \addplot[color=purple,mark=square,mark size=1.3pt]
                    coordinates{
                     (1, 43.4)(5, 43.2)(10, 43.4)(30, 43.6)(50, 43.7)(70, 43.5)
                    };\label{plots:MSANet}
            \nextgroupplot[title={Res-101, Fold4}]
            \addplot[color=blue,mark=square,mark size=1.3pt]
                    coordinates{
                     (1,41.55)(5,48.60)(10,48.49)(30,48.89)(50, 48.60)(70, 47.26)
                    };\label{plots:DCAMA}
             \addplot[color=orange,mark=square,mark size=1.3pt]
                    coordinates{
                     (1,40.31)(5,50.89)(10,53.52)(30,57.31)(50, 57.92)(70, 57.39)
                    };\label{plots:Ours}
            \addplot[color=purple,mark=square,mark size=1.3pt]
                    coordinates{
                     (1, 50.0)(5, 53.6)(10, 47.1)(30, 52.0)(50, 48.9)(70, 49.8)
                    };\label{plots:MSANet}
            \nextgroupplot[title={Swin, Fold4}]
            \addplot[color=blue,mark=square,mark size=1.3pt]
                    coordinates{
                     (1,47.75)(5,57.78)(10,59.98)(30,61.82)(50, 61.90)(70, 62.99)
                    };\label{plots:DCAMA}
            \addplot[color=orange,mark=square,mark size=1.3pt]
                    coordinates{
                     (1,48.43)(5,57.88)(10,60.92)(30,63.20)(50, 64.54)(70, 64.76)
                    };\label{plots:Ours}
           \nextgroupplot[title={Res-50, Fold1},ylabel={mIoU, PASCAL}]
            \addplot[color=blue,mark=square,mark size=1.3pt]
                    coordinates{
                     (1,67.42)(5,71.68)(10,71.53)(30, 68.15)(50, 65.63)(70, 64.49)
                    };\label{plots:DCAMA}
            \addplot[color=orange,mark=square,mark size=1.3pt]
                    coordinates{
                     (1,66.75)(5,72.06)(10,72.66)(30, 73.93)(50, 74.12)(70, 74.48)
                    };\label{plots:Ours}
            \addplot[color=purple,mark=square,mark size=1.3pt]
                    coordinates{
                     (1,63.0)(5, 62.4)(10, 64.4)(30, 61.2)(50, 62.7)(70, 63.4)
                    };\label{plots:MSANet}
            \coordinate (top) at (rel axis cs:0,1);
            \nextgroupplot[title={Res-101, Fold1}]
            \addplot[color=blue,mark=square,mark size=1.3pt]
                    coordinates{
                     (1,65.43)(5,69.31)(10,71.62)(30,70.16)(50, 67.66)(70, 67.82)
                    };\label{plots:DCAMA}
            \addplot[color=orange,mark=square,mark size=1.3pt]
                    coordinates{
                     (1,66.93)(5,73.11)(10,75.46)(30, 77.21)(50, 77.67)(70, 77.97)
                    };\label{plots:Ours}
            \addplot[color=purple,mark=square,mark size=1.3pt]
                    coordinates{
                     (1,68.0)(5, 67.8)(10, 71.0)(30, 66.5)(50, 67.7)(70, 67.9)
                    };\label{plots:MSANet}
           \nextgroupplot[title={Swin, Fold1}]
            \addplot[color=blue,mark=square,mark size=1.3pt]
                    coordinates{
                     (1,72.02)(5,74.05)(10,76.41)(30,77.95)(50, 77.39)(70, 77.39)
                    };\label{plots:DCAMA}
            \addplot[color=orange,mark=square,mark size=1.3pt]
                    coordinates{
                     (1,69.06)(5,75.61)(10,77.20)(30,78.14)(50, 77.91)(70, 79.12)
                    };\label{plots:Ours}
            \nextgroupplot[title={Res-50, Fold2}]
            \addplot[color=blue,mark=square,mark size=1.3pt]
                    coordinates{
                     (1,72.31)(5,74.18)(10,74.11)(30,71.06)(50, 69.40)(70, 67.97)
                    };\label{plots:DCAMA}
            \addplot[color=orange,mark=square,mark size=1.3pt]
                    coordinates{
                     (1,71.37)(5,74.69)(10,74.80)(30,74.94)(50, 75.4)(70, 76.14)
                    };\label{plots:Ours}
            \addplot[color=purple,mark=square,mark size=1.3pt]
                    coordinates{
                     (1,71.0)(5, 74.9)(10, 70.6)(30, 71.2)(50, 70.9)(70, 70.5)
                    };\label{plots:MSANet}
            \nextgroupplot[title={Res-101, Fold2}]
            \addplot[color=blue,mark=square,mark size=1.3pt]
                    coordinates{
                     (1,71.49)(5,72.86)(10,71.82)(30, 75.38)(50, 75.60)(70, 76.25)
                    };\label{plots:DCAMA}
             \addplot[color=purple,mark=square,mark size=1.3pt]
                    coordinates{
                     (1,72.3)(5, 73.2)(10, 75.6)(30, 75.3)(50, 72.7)(70, 73.2)
                    };\label{plots:MSANet}
            \addplot[color=orange,mark=square,mark size=1.3pt]
                    coordinates{
                     (1,70.71)(5,73.75)(10,75.20)(30, 77.55)(50,76.98)(70,78.08)
                    };\label{plots:Ours}
            \nextgroupplot[title={Swin, Fold2}]
             \addplot[color=blue,mark=square,mark size=1.3pt]
                    coordinates{
                     (1,73.85)(5,76.20)(10,75.59)(30,78.00)(50, 78.33)(70, 78.54)
                    };\label{plots:DCAMA}
            \addplot[color=orange,mark=square,mark size=1.3pt]
                    coordinates{
                     (1,75.06)(5,76.78)(10,76.97)(30,78.02)(50, 80.51)(70, 81.86)
                    };\label{plots:Ours}
            \nextgroupplot[title={Res-50, Fold3}, ylabel={mIoU, PASCAL}]
            \addplot[color=blue,mark=square,mark size=1.3pt]
                    coordinates{
                     (1,59.00)(5,63.62)(10,63.89)(30,60.80)(50, 58.56)(70, 56.97)
                    };\label{plots:DCAMA}
            \addplot[color=orange,mark=square,mark size=1.3pt]
                    coordinates{
                     (1,58.36)(5,63.40)(10,72.16)(30,72.38)(50,75.15)(70,75.23)
                    };\label{plots:Ours}
             \addplot[color=purple,mark=square,mark size=1.3pt]
                    coordinates{
                     (1,67.8)(5, 64.2)(10, 64.0)(30, 65.3)(50, 65.4)(70, 65.2)
                    };\label{plots:MSANet}
            \nextgroupplot[title={Res-101, Fold3}]
             \addplot[color=blue,mark=square,mark size=1.3pt]
                    coordinates{
                     (1,65.49)(5,67.43)(10,64.93)(30,58.48)(50, 53.93)(70, 51.03)
                    };\label{plots:DCAMA}
            \addplot[color=orange,mark=square,mark size=1.3pt]
                    coordinates{
                     (1,68.67)(5,73.09)(10,73.85)(30,74.27)(50,74.61)(70,74.66)
                    };\label{plots:Ours}
            \addplot[color=purple,mark=square,mark size=1.3pt]
                    coordinates{
                     (1, 64.7)(5, 65.7)(10, 66.0)(30, 67.4)(50, 65.5)(70, 64.7)
                    };\label{plots:MSANet}
           \nextgroupplot[title={Swin, Fold3}]
            \addplot[color=blue,mark=square,mark size=1.3pt]
                    coordinates{
                     (1,65.05)(5,72.22)(10,74.11)(30, 75.38)(50, 76.09)(70, 75.59)
                    };\label{plots:DCAMA}
            \addplot[color=orange,mark=square,mark size=1.3pt]
                    coordinates{
                     (1,69.37)(5,77.10)(10,79.37)(30, 81.85)(50, 82.51)(70, 82.58)
                    };\label{plots:Ours}
            \nextgroupplot[title={Res-50, Fold4}, ]
             \addplot[color=blue,mark=square,mark size=1.3pt]
                    coordinates{
                     (1,59.00)(5,64.79)(10,66.16)(30,64.95)(50, 63.39)(70, 62.92)
                    };\label{plots:DCAMA}
            \addplot[color=orange,mark=square,mark size=1.3pt]
                    coordinates{
                     (1,56.21)(5,68.69)(10,71.92)(30,72.41)(50,73.92)(70,73.84)
                    };\label{plots:Ours}
             \addplot[color=purple,mark=square,mark size=1.3pt]
                    coordinates{
                     (1,61.7)(5, 60.0)(10, 59.7)(30, 59.8)(50, 59.4)(70, 59.8)
                    };\label{plots:MSANet}
            \nextgroupplot[title={Res-101, Fold4}]
            \addplot[color=blue,mark=square,mark size=1.3pt]
                    coordinates{
                     (1,57.57)(5,65.77)(10,68.81)(30,70.65)(50, 70.83)(70, 70.61)
                    };\label{plots:DCAMA}
             \addplot[color=orange,mark=square,mark size=1.3pt]
                    coordinates{
                     (1,57.98)(5,67.61)(10,72.36)(30,74.68)(50, 75.08)(70, 75.44)
                    };\label{plots:Ours}
            \addplot[color=purple,mark=square,mark size=1.3pt]
                    coordinates{
                     (1, 61.6)(5, 63.1)(10, 61.5)(30, 62.4)(50, 62.6)(70, 62.1)
                    };\label{plots:MSANet}
            \nextgroupplot[title={Swin, Fold4}]
            \addplot[color=blue,mark=square,mark size=1.3pt]
                    coordinates{
                     (1,66.66)(5,71.08)(10,76.85)(30,77.56)(50, 77.85)(70, 77.75)
                    };\label{plots:DCAMA}
            \addplot[color=orange,mark=square,mark size=1.3pt]
                    coordinates{
                     (1,65.19)(5,74.18)(10,77.00)(30, 78.48)(50, 78.76)(70, 78.63)
                    };\label{plots:Ours}
            
        \coordinate (bot) at (rel axis cs:1,0);
            \end{groupplot}
              \path (top|-current bounding box.north)--
                    coordinate(legendpos)
                    (bot|-current bounding box.north);
              \matrix[
                  matrix of nodes,
                  anchor=south,
                  draw,
                  inner sep=0.2em,
                  draw
                ]at([yshift=1ex]legendpos)
                {
                  \ref{plots:DCAMA}& DCAMA &[5pt]

                  \ref{plots:MSANet} & MSANet&[5pt]
                  
                  \ref{plots:Ours}& Ours
                  
                \\};
        \end{tikzpicture}
    \caption{We compare the segmentation mIoU metric with MSANet and DCAMA. We report results on two FSS benchmarks, COCO-20$^i$ and PASCAL-5$^i$, both FSS benchmarks have four data folds and we implement three backbones, hence we totally illustrate 24 subplots.}
    \label{fig:result-coco}
\end{figure*}
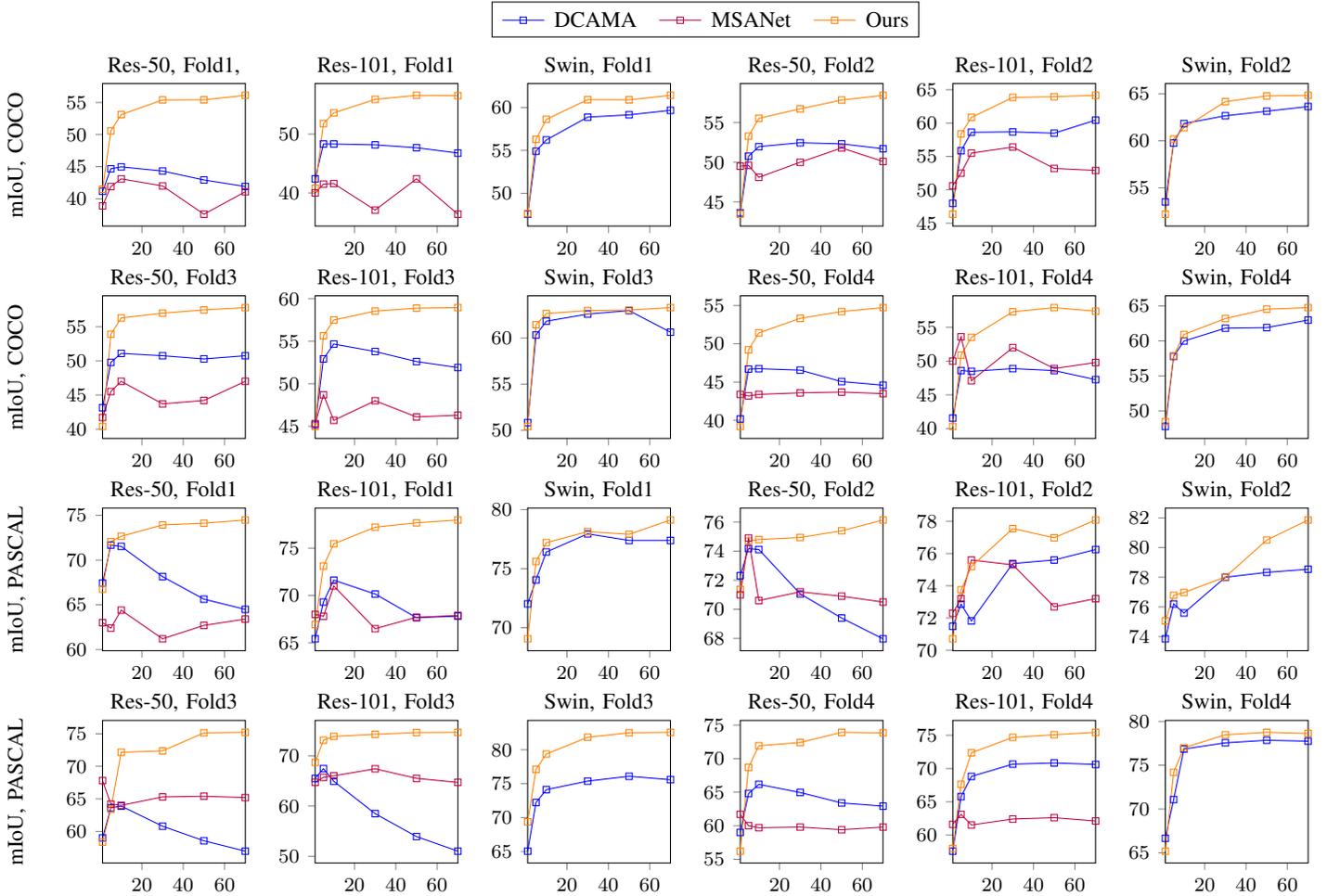
\begin{table} [htp]
\caption{Evaluations on COCO-20$^i$ and PASCAL-5$^i$ follow the 4-fold cross-validation. The statistics include the per-fold testing category names (indexes) as well as the per-category image numbers. For each fold, the training/testing category numbers of COCO-20$^i$ and PASCAL-5$^i$ are 20/60 and 5/15.}
\label{tab: statistics}
\centering
    \begin{subtable}{\textwidth}
        \setlength{\tabcolsep}{0.45pt}
\begin{footnotesize}
\begin{tabular}{rl|rl|rl|rl}
\hline 
\multicolumn{2}{c|}{COCO-$20^1$} & \multicolumn{2}{c|}{COCO-$20^2$} & \multicolumn{2}{c|}{COCO-$20^3$} & \multicolumn{2}{c}{COCO-$20^4$} \\ \hline
1       & Person(19217)            & 2      & Bicycle(748)           & 3      & Car(2754)               & 4      & Motorcycle(1073)        \\
5       & Airplane(727)          & 6      & Bus(1213)               & 7      & Train(1257)             & 8      & Truck(1565)             \\
9       & Boat(816)              & 10     & T.light(572)     & 11     & Fire H.(411)      & 12     & Stop(385)          \\
13      & Park meter(172)     & 14     & Bench(1379)             & 15     & Bird(713)              & 16     & Cat(1291)               \\
17      & Dog(1203)               & 18     & Horse(925)             & 19     & Sheep(456)             & 20     & Cow(616)               \\
21      & Elephant(699)          & 22     & Bear(335)              & 23     & Zebra(655)             & 24     & Giraffe(839)           \\
25      & Backpack(805)          & 26     & Umbrella(1135)          & 27     & Handbag(998)           & 28     & Tie(154)               \\
29      & Suitcase(709)          & 30     & Frisbee(270)           & 31     & Skis(434)              & 32     & Snowboard(272)         \\
33      & Sports ball(142)       & 34     & Kite(404)              & 35     & B. bat(173)      & 36     & B. glove(159)    \\
37      & Skateboard(655)        & 38     & Surfboard(804)         & 39     & T. racket(687)     & 40     & Bottle(1351)            \\
41      & W. glass(516)        & 42     & Cup(1733)               & 43     & Fork(419)              & 44     & Knife(470)             \\
45      & Spoon(312)             & 46     & Bowl(1577)              & 47     & Banana(551)            & 48     & Apple(298)             \\
49      & Sandwich(577)          & 50     & Orange(380)            & 51     & Broccoli(521)          & 52     & Carrot(397)            \\
53      & Hot dog(331)           & 54     & Pizza(916)             & 55     & Donut(423)             & 56     & Cake(790)              \\
57      & Chair(3648)             & 58     & Couch(1375)             & 59     & P. plant(1198)      & 60     & Bed(1272)               \\
61     & D. table(3722)      & 62     & Toilet(1116)            & 63     & TV(1416)                & 64     & Laptop(1070)            \\
65      & Mouse(200)             & 66     & Remote(302)            & 67     & Keyboard(619)          & 68     & Cellphone(446)        \\
69      & Microwave(389)         & 70     & Oven(925)              & 71     & Toaster(38)           & 72     & Sink(1097)              \\
73      & Fridge(773)      & 74     & Book(1159)              & 75     & Clock(803)             & 76     & Vase(676)              \\
77      & Scissors(170)          & 78     & Teddy(644)        & 79     & Hairdrier(37)        & 80     & Toothbrush(94)     \\
\hline
\end{tabular}
\end{footnotesize}
    \end{subtable}

    \centering
    \begin{subtable}{\textwidth}
    \vspace{8pt}
        \setlength{\tabcolsep}{0.45pt}
\begin{footnotesize}
\begin{tabular}{rl|rl|rl|rl}
\hline 
\multicolumn{2}{c|}{PASCAL-$5^1$} & \multicolumn{2}{c|}{PASCAL-$5^2$} & \multicolumn{2}{c|}{PASCAL-$5^3$} & \multicolumn{2}{c}{PASCAL-$5^4$} \\ \hline
1       & Aeroplane(670)            & 2      & Bicycle(552)           & 3      & Bird(765)               & 4      & Boat(508)        \\
5       & Bottle(706)          & 6      & Bus(421)               & 7      & Car(1161)             & 8      & Cat(1080)             \\
9       & Chair(1119)              & 10     & Cow(303)     & 11     & Diningtable(538)      & 12     & Dog(1286)          \\
13      & Horse(482)     & 14     & Motorbike(526)             & 15     & Person(4087)              & 16     & Pottedplant(527)               \\
17      & Sheep(325)               & 18     & Sofa(507)             & 19     & Train(544)             & 20     & TVmonitor(575)               \\
\hline
\end{tabular}
\end{footnotesize}
    \end{subtable}
\end{table}
%
%

%
In Sec.~\ref{exp: cross-domain exp}, to verify the robustness of our method, we use COCO-20$^{i}$ and PASCAL-5$^i$ to conduct cross-domain tests on their shared 17 categories. 
We train models on COCO-20$^{i}$ then apply the model for testing queries from PASCAL-5$^i$. This experiment can validate if our network can generalize well against the data domain gap.
%

%
In Sec.~\ref{exp: plug-and-play exp}, to verify the plug-in ability of our solution, we add the technical contributions (Sec.~\ref{method SC} and Sec.~\ref{method SIP}) to two SOTA FSS methods~\cite{iqbal2022msanet, peng2023hierarchical}. 
Quantitative and qualitative results on COCO-20$^{i}$ and PASCAL-5$^i$ demonstrate that our techniques can support existing FSS approaches against support dilution with a simple plug-in operation. 

In Sec.~\ref{exp: online}, we apply our method for online semantic segmentation to verify its practical superiority. 
Our model is trained on COCO-20$^{i}$. At the inference stage, for each COCO-20$^{i}$ query, as well as its certain category name, we use the name as the keyword to collect related images through Google search, then use the top $N$ relevant images to build our support set. The corresponding support masks are obtained by the widely-used large vision model Grounding-SAM~\cite{ren2401grounded}; we also analyze the effects of the noisy support masks. 

In Sec.~\ref{exp: ablation}, we conduct experiments on COCO-20$^i$ for ablation studies and test-time analysis. 
%

%
In Sec.~\ref{exp: real-world}, we conduct two experiments, on indoor and autonomous driving benchmarks respectively, to show that our method has strong potential for real-world applications. 
More than benchmarked scenarios, we capture daily real images with a mobile phone as queries. Our method is easy to deploy on the device and achieves reliable real-time results. 

Please read the following sections for more details.

\subsection{Implementation Details}
\label{exp: imple}
Our method is implemented with the PyTorch~\cite{paszke2019pytorch} framework.  
When training, to speed up the time for convergence, we partly initialize the pipeline (Sec.~\ref{method pipeline}) with the checkpoint provided by DCAMA. 
Specifically, for the Backbone and the Refiner, we initialize it with exactly the same parameters of DCAMA.
For our SC modules (Eq.~\ref{SymAtt3}), we use the parameters of the corresponding DCAMA's query FFN to initialize $f_1^l$, and initialize all $f_2^l$s' weight matrices to be zeros and the biases to be ones (i.e., we assume all the support features have identical objectness).
Note that our Support Image Pruning, as a retrieval operation, is not involved in the backward propogation. When inference, if the number of shots $N$ is larger than 30, we will apply Support Image Pruning and keep only 30 support images for the sequential SCs (i.e., $N'=30$). 
We implement our model with three different backbones, ResNet-50, ResNet-101 and Swin-Transformer-Base. The input sizes of both support and query images are $384\times384$. We use the SGD optimizer for training, the learning rate, momentum, and weight decay are initialized to 0.0001, 0.9, and 0.0001, respectively. For fair comparisons, we use no data augmentation, exactly following previous methods such as DCAMA~\cite{shi2022dense}. The model is trained for 5 epochs on a TiTAN XP GPU and the total training process costs less than an hour. We will release the complete code upon paper acceptance.

\subsection{$N$-shot Comparison Experiments}
\label{exp: normal exp}
We adopt the SOTA FSS methods DCAMA~\cite{shi2022dense} and 
MSANet~\cite{iqbal2022msanet} for comparison. Here, we briefly introduce how the two works deal with multiple shots. 
MSANet processes $N$ support images individually, it generates a correlation map between each support image and each query image for $N$ times then takes the average correlation for query mask decoding. 
%
%
Unfortunately, due to the independent correlation calculation, MSANet cannot fully explore the inter-support correlation. 
DCAMA takes a step forward, it enables the cross-attention module to process all the support features simultaneously (Eq.~\ref{Asq2}), so it can model the inter-support correlation. 
However, when the number of supports gets larger, DCAMA struggles to concentrate on the high-contributed supports, and the support dilution problem impairs the segmentation performance.

\begin{figure*}[t]
    \centering
\includegraphics[width=0.97\textwidth]{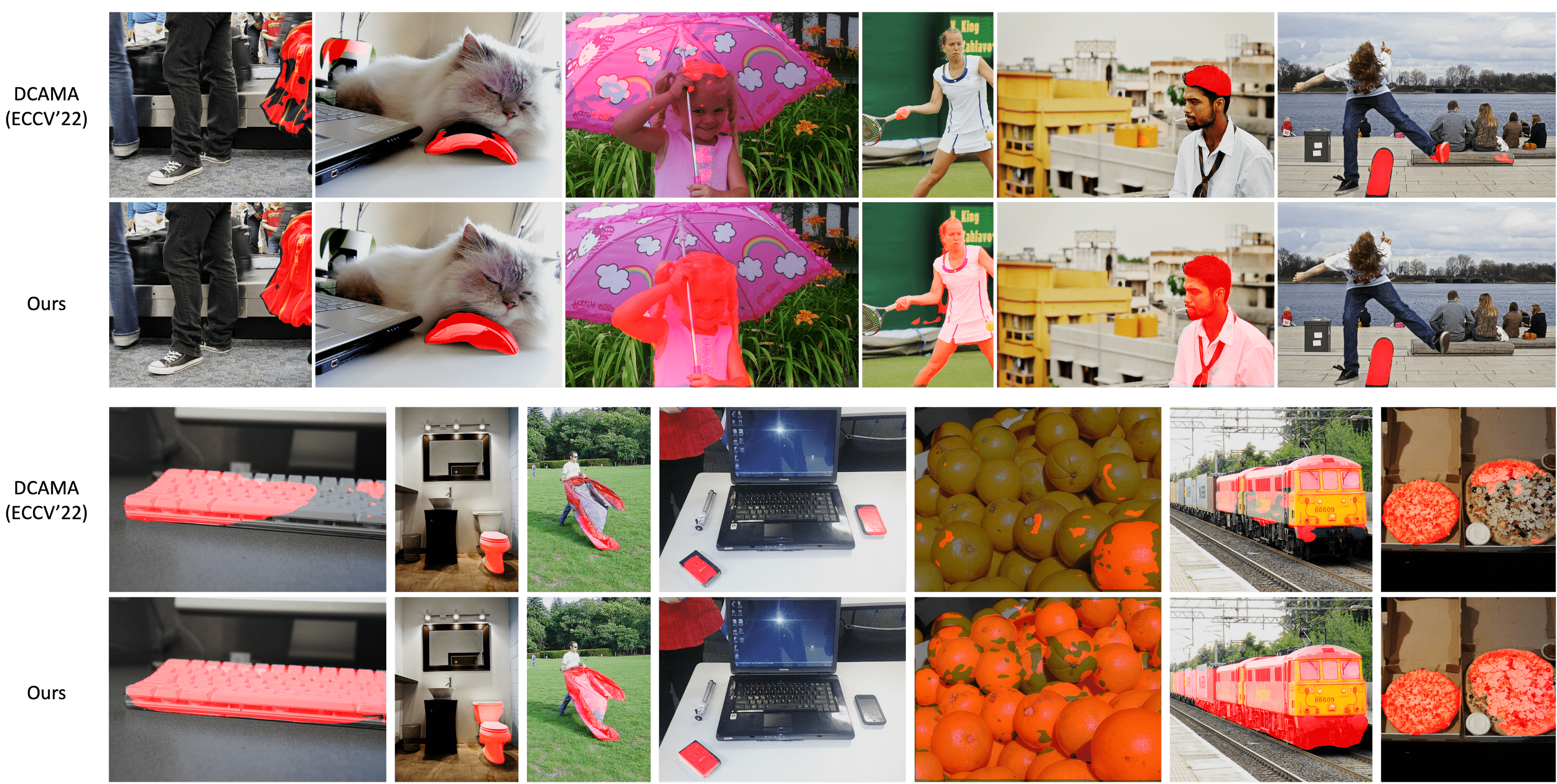}
    \caption{Qualitative comparison with DCAMA on COCO-20$^i$, $N=20$, backbone is Swin-Transfromer-Base. Given more supports (accompanied by support dilution), our method significantly outperforms DCAMA, we produce complete and accurate masks for novel-category objects.}
    \label{fig: Comp}
\end{figure*}

We report the quantitative segmentation results on COCO-20$^i$ and PASCAL-5$^i$ in Fig.~\ref{fig:result-coco}.
Our method achieves the best performance for both benchmarks, across different data folds and different network backbones (following the original settings of the two papers, for MSANet, we implement Res-50 and Res-101; for DCAMA, we implement Res-50, Res-101, and Swin-B). 
Particularly, there are two observations worth noticing.
First, our method outperforms others when there is only one support image (i.e., 1-shot FSS). Shown in Fig.~\ref{fig:result-coco}, in most subplots, ours can start at a higher initial point, which means that SC can better model one-to-one support-query correlation.
Second, as $N$ gets larger, the mask precision of other methods gains slow improvement or even drops, on the contrary, our mIoU value keeps climbing from 1 to 70 shots. Given a large support pool, our method can neglect the low-contributed information and consistently concentrate on those high-contributed supports, indicating that SC can better model multi-to-one support-query correlation.
Typical qualitative results on COCO-20$^i$ can be found in Fig.~\ref{fig: Comp}.
%
%
\subsection{Cross-Domain Comparison Experiments}
\label{exp: cross-domain exp}
\begin{figure*}[t]
    \centering
\includegraphics[width=1\textwidth]{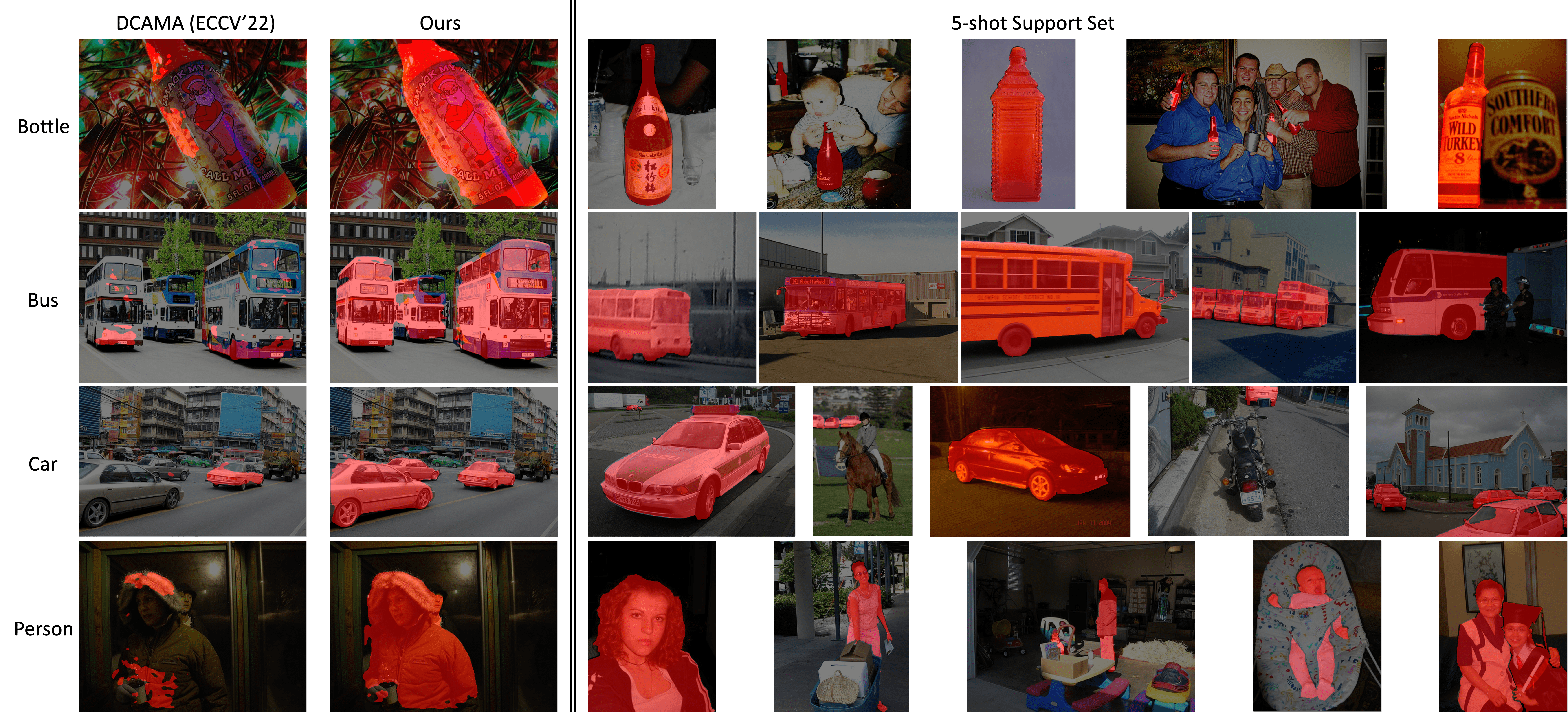}
    \caption{Typical results on PASCAL-5$^i$. Note that the training support-query pairs are from COCO-20$^i$ while the testing support-query pairs are from PASCAL-5$^i$, $N=5$, backbone is Swin-Transformer-Base. Thanks to Symmetric Correlation (SC), our method is robust to data distribution bias and shows significantly better cross-domain FSS results than DCAMA.}
    \label{fig:Cross Domain}
\end{figure*}

\begin{figure}[t]
    \centering
\includegraphics[width=\linewidth]{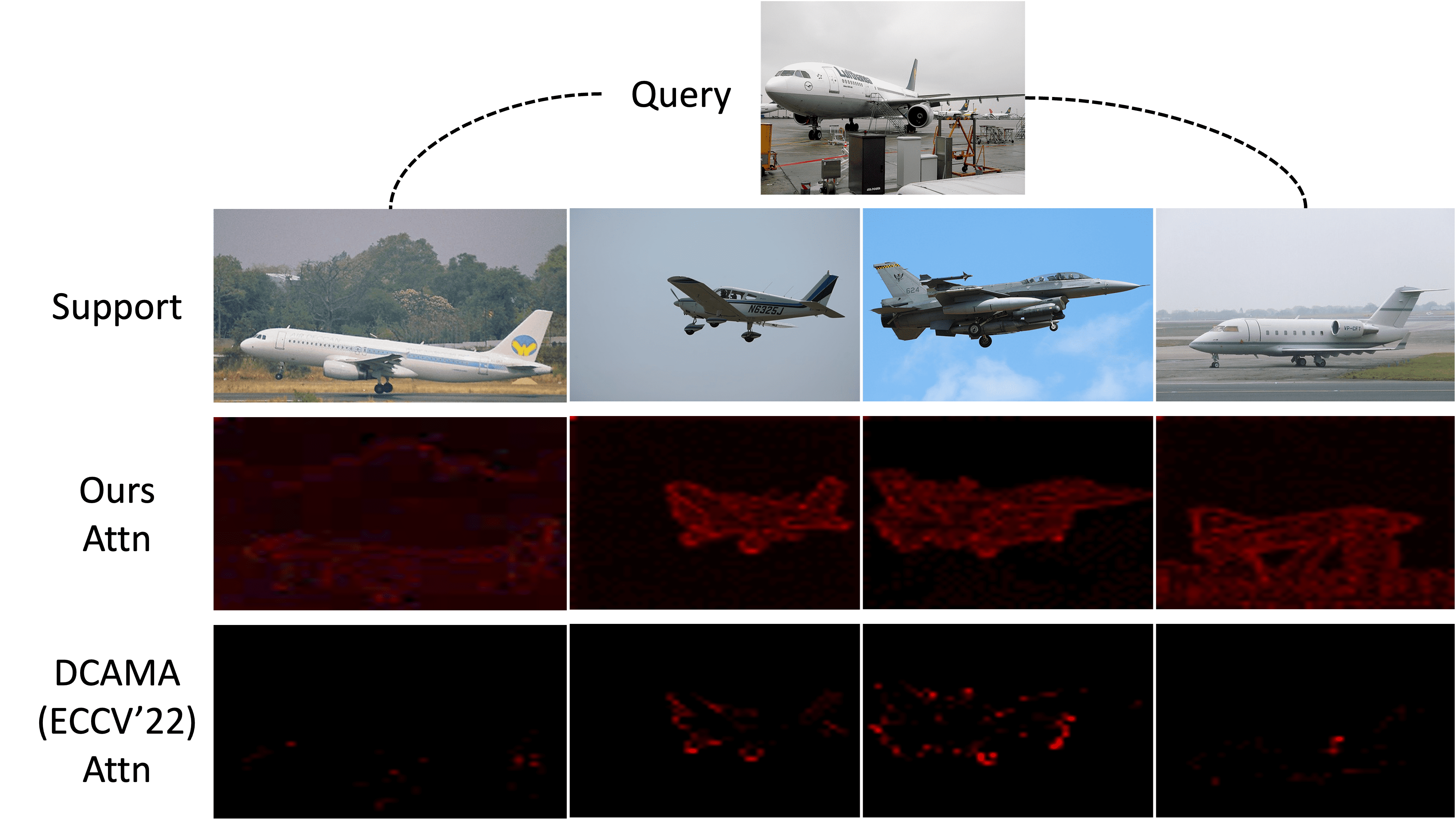}
    \caption{Illustration of support-query correlation heatmaps in the cross-domain experiment, where our method can successfully concentrate the object regions. $N=30$, we illustrate four examples.}
    \label{fig:Cross Domain Attn}
\end{figure}

\begin{table*}[t]
\centering
\caption{mIou results of cross-domain experiments, the training pairs are from COCO-20$^i$ and the testing pairs are from PASCAL-5$^i$.}
\centering
\label{tab: cross domain1}
\centering
\resizebox{\textwidth}{!}{
\begin{tabular}{c||cccccl|cccccl|cccccl}
\hline
Backbone & \multicolumn{6}{c||}{ResNet-50} & \multicolumn{6}{c||}{ResNet-101} & \multicolumn{6}{c}{Swin-Transformer-Base} \\
 \hline
     shots & \multicolumn{1}{c|}{1} & \multicolumn{1}{c|}{5} & \multicolumn{1}{c|}{10} & \multicolumn{1}{c|}{30} & \multicolumn{1}{c|}{50} & \multicolumn{1}{c||}{70}
        & \multicolumn{1}{c|}{1} & \multicolumn{1}{c|}{5} & \multicolumn{1}{c|}{10} & \multicolumn{1}{c|}{30} & \multicolumn{1}{c|}{50} & \multicolumn{1}{c||}{70}
        & \multicolumn{1}{c|}{1} & \multicolumn{1}{c|}{5} & \multicolumn{1}{c|}{10} & \multicolumn{1}{c|}{30} & \multicolumn{1}{c|}{50} & 70
\\ 
\hline
DCAMA   & \multicolumn{1}{c|}{48.5}  & \multicolumn{1}{c|}{46.1}  & \multicolumn{1}{c|}{44.6}   & \multicolumn{1}{c|}{50.9}   & \multicolumn{1}{c|}{51.3} &\multicolumn{1}{c||}{50.1}   &
        \multicolumn{1}{c|}{49.0}  & \multicolumn{1}{c|}{48.9}  & \multicolumn{1}{c|}{48.1}   & \multicolumn{1}{c|}{45.0}   & \multicolumn{1}{c|}{46.6}   &  \multicolumn{1}{c||}{49.1} & 
        \multicolumn{1}{c|}{44.4}  & \multicolumn{1}{c|}{47.3}  & \multicolumn{1}{c|}{46.8}   & \multicolumn{1}{c|}{45.3}   & \multicolumn{1}{c|}{46.1} & \multicolumn{1}{c}{46.4} 
\\ 
\hline

 Ours    & \multicolumn{1}{c|}{\textbf{61.5}}  & \multicolumn{1}{c|}{\textbf{62.3}}  & \multicolumn{1}{c|}{\textbf{63.9}}   & \multicolumn{1}{c|}{\textbf{62.4}}   & \multicolumn{1}{c|}{\textbf{63.1}}   &  \multicolumn{1}{c||}{\textbf{63.5}}
         & \multicolumn{1}{c|}{\textbf{67.1}}  & \multicolumn{1}{c|}{\textbf{74.2}}  & \multicolumn{1}{c|}{\textbf{74.0}}   & \multicolumn{1}{c|}{\textbf{73.9}}   & \multicolumn{1}{c|}{\textbf{76.2}} 
          & \multicolumn{1}{c||}{\textbf{76.5}}  & \multicolumn{1}{c|}{\textbf{58.8}}  & \multicolumn{1}{c|}{\textbf{62.3}}   & \multicolumn{1}{c|}{\textbf{63.3}}   & \multicolumn{1}{c|}{\textbf{63.1}} & \multicolumn{1}{c|}{\textbf{68.4}} & {\textbf{68.6}} 
\\ \hline
\end{tabular}}
\end{table*}
\begin{table}[t]
\centering
\caption{mIoU(\%) results of plug-and-play experiments (Top: COCO-20$^i$; bottom: PASCAL-5$^i$.) * indicates adding our SC and pruning to the original method. }
\label{tab: plug}
\resizebox{0.45\textwidth}{!}{
\begin{tabular}{c||cccccc}
\hline
\multirow{2}{*}{Methods} & \multicolumn{5}{c}{shots}
        \\ \cline{2-7}
        & \multicolumn{1}{c|}{1}     & \multicolumn{1}{c|}{5}     & \multicolumn{1}{c|}{10}    & \multicolumn{1}{c|}{30}    & \multicolumn{1}{c|}{50} & 70    \\ \hline
HDMNet  & \multicolumn{1}{c|}{40.70} & \multicolumn{1}{c|}{46.18} & \multicolumn{1}{c|}{47.37} & \multicolumn{1}{c|}{46.66} & \multicolumn{1}{c|}{47.86} & 47.19     \\ \hline
HDMNet*  & \multicolumn{1}{c|}{46.20} & \multicolumn{1}{c|}{48.32} & \multicolumn{1}{c|}{49.52} & \multicolumn{1}{c|}{52.12} & \multicolumn{1}{c|}{53.44} & 53.51     \\ \hline
MSANet   & \multicolumn{1}{c|}{38.88} & \multicolumn{1}{c|}{41.89} & \multicolumn{1}{c|}{43.05} & \multicolumn{1}{c|}{41.97} & \multicolumn{1}{c|}{37.55} & 41.12 \\ \hline
MSANet*   & \multicolumn{1}{c|}{41.33} & \multicolumn{1}{c|}{44.83} & \multicolumn{1}{c|}{46.18} & \multicolumn{1}{c|}{47.21} & \multicolumn{1}{c|}{47.35} & 47.39 \\ \hline
DCAMA   & \multicolumn{1}{c|}{41.17} & \multicolumn{1}{c|}{44.67} & \multicolumn{1}{c|}{44.97} & \multicolumn{1}{c|}{44.32} & \multicolumn{1}{c|}{42.93} & 41.91 \\ \hline
DCAMA*   & \multicolumn{1}{c|}{41.49} & \multicolumn{1}{c|}{50.58} & \multicolumn{1}{c|}{53.11} & \multicolumn{1}{c|}{55.40} & \multicolumn{1}{c|}{55.42} & 56.11 \\ \hline

\hline
HDMNet  & \multicolumn{1}{c|}{71.14} & \multicolumn{1}{c|}{71.26} & \multicolumn{1}{c|}{72.44} & \multicolumn{1}{c|}{71.92} & \multicolumn{1}{c|}{71.98} & 72.51     \\ \hline
HDMNet*  & \multicolumn{1}{c|}{72.98} & \multicolumn{1}{c|}{73.24} & \multicolumn{1}{c|}{73.55} & \multicolumn{1}{c|}{73.49} & \multicolumn{1}{c|}{74.10} & 74.13     \\ \hline
MSANet   & \multicolumn{1}{c|}{63.02} & \multicolumn{1}{c|}{62.41} & \multicolumn{1}{c|}{64.08} & \multicolumn{1}{c|}{61.19} & \multicolumn{1}{c|}{62.66} & 63.42  \\ \hline
MSANet*   & \multicolumn{1}{c|}{63.56} & \multicolumn{1}{c|}{64.28} & \multicolumn{1}{c|}{64.97} & \multicolumn{1}{c|}{65.01} & \multicolumn{1}{c|}{65.00} & 65.15 \\ \hline
DCAMA   & \multicolumn{1}{c|}{67.42} & \multicolumn{1}{c|}{71.68} & \multicolumn{1}{c|}{71.53} & \multicolumn{1}{c|}{68.15} & \multicolumn{1}{c|}{65.63} & 64.49  \\ \hline
DCAMA*   & \multicolumn{1}{c|}{66.75} & \multicolumn{1}{c|}{72.06} & \multicolumn{1}{c|}{72.66} & \multicolumn{1}{c|}{73.93} & \multicolumn{1}{c|}{74.12} & 74.48  \\ \hline                
\end{tabular}}
\end{table}
To measure the model's generalization ability, we train the network on COCO-20$^i$ and then conduct inference on PASCAL-5$^i$.
%
%
The quantitative results on the shared 17 categories are reported in Tab.~\ref{tab: cross domain1}, and the typical qualitative results are shown in Fig.~\ref{fig:Cross Domain}.
Compared with DCAMA(Swin-B), our method gains significant cross-domain segmentation improvement. 
The reason is that, existing FSS correlation modules highly depend on the feature quality, once the image feature suffers from distribution bias, the correlation results are less reliable. 
On the contrary, SC is designed to understand the \textit{relative} contributions between multiple supports, it drives the problem from `determine if a support is important' to `distinguish which support is more important', this ability enhances the network's robustness against the data distribution gap. In Fig.~\ref{fig:Cross Domain Attn}, we illustrate the correlation heatmaps as proof. Given multiple supports, SC can describe clear object regions for all of them, while DCAMA loses objectness awareness due to support dilution. 

\subsection{Plug-and-play Experiments}
\label{exp: plug-and-play exp}
Our Symmetric Correlation (Sec.~\ref{method SC}) and the Support Image Pruning operation (Sec.~\ref{method SIP}) can be simple plug-in modules to enhance many FSS methods against support dilution. 
We conduct plug-and-paly experiments for three SOTA FSS methods HDMNet~\cite{min2021hypercorrelation}, MSANet~\cite{iqbal2022msanet} and DCAMA~\cite{shi2022dense}. 
DCAMA(Swin-B) has a similar architecture to our method, we directly replace their correlation module with our SC and add the pruning operation. 
For HDMNet and MSANet, we concatenate the input supports $\{I_{S_i} \in \mathbb{R}^{H \times W \times 3}\}_{i=1}^N$ to an image sequence $I_S \in \mathbb{R}^{NH \times W \times 3}$, then we feed $I_S$ into the backbone, directly followed by the Support Image Pruning operation. For feature-level correlation learning, we replace the correlation map and the similarity module with Eq.~\ref{SymAtt2} for HDMNet and MSANet respectively. 
Shown in Tab.~\ref{tab: plug}, equipped with our designs, FSS methods become robust against support dilution and gain considerable segmentation improvements, demonstrating that our method is a general solution and has wide prospects for general usage.
\subsection{Online Demonstration Experiments}
\label{exp: online}
\begin{figure}
    \centering   \includegraphics[width=0.5\textwidth]{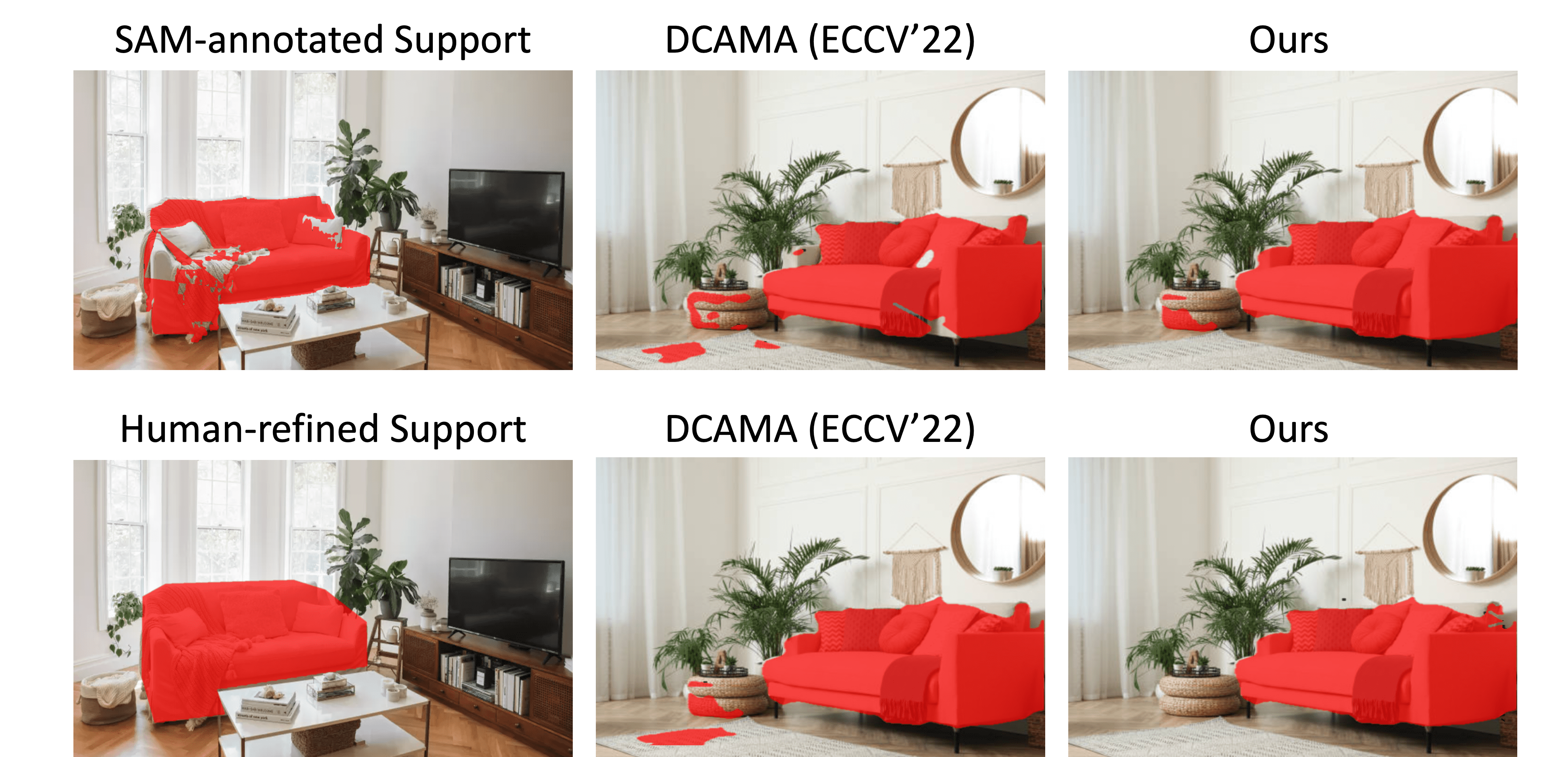}
    \caption{Mask results of using noisy support masks generated by Grounding-SAM (top) v.s. using perfect masks from human annotator (bottom). The query is from COCO-20$^i$ and the supports are searched from Google.}
    \label{fig: noise}
\end{figure}
\begin{table}[t]
\centering
\caption{Online segmentation mIoU(\%) results. Support images are from Google. * indicates using human-corrected support masks.}
\label{tab: llm-comp}
\resizebox{0.45\textwidth}{!}{
\begin{tabular}{c||cccccc}
\hline
\multirow{2}{*}{Methods} & \multicolumn{5}{c}{shots}
        \\ \cline{2-7}
        & \multicolumn{1}{c|}{1}     & \multicolumn{1}{c|}{5}     & \multicolumn{1}{c|}{10}    & \multicolumn{1}{c|}{30}    & \multicolumn{1}{c|}{50} & 70    \\ \hline
DCAMA & \multicolumn{1}{c|}{15.31} & \multicolumn{1}{c|}{18.04} & \multicolumn{1}{c|}{21.88} & \multicolumn{1}{c|}{21.93} & \multicolumn{1}{c|}{24.61} & \multicolumn{1}{c}{24.79}     \\ \hline
DCAMA*  & \multicolumn{1}{c|}{21.43} & \multicolumn{1}{c|}{23.54} & \multicolumn{1}{c|}{26.77} & \multicolumn{1}{c|}{26.48} & \multicolumn{1}{c|}{26.93} & 27.60     \\ \hline
Ours   & \multicolumn{1}{c|}{23.14} & \multicolumn{1}{c|}{30.68} & \multicolumn{1}{c|}{34.96} & \multicolumn{1}{c|}{36.71} & \multicolumn{1}{c|}{38.43} & \multicolumn{1}{c}{39.26} \\ \hline
Ours*   & \multicolumn{1}{c|}{\textbf{25.63}} & \multicolumn{1}{c|}{\textbf{32.47}} & \multicolumn{1}{c|}{\textbf{35.09}} & \multicolumn{1}{c|}{\textbf{37.66}} & \multicolumn{1}{c|}{\textbf{39.83}} & \textbf{40.21}  \\ \hline
\end{tabular}}
\end{table}
We extend benchmarked FSS to online FSS, which is no longer limited by the human-compiled support data, hence is more helpful in the real world.
In practice, we first pretrain the model on COCO-20$^i$ training split. Then, given a COCO-20$^i$ testing split query image, as well as its category name as the keyword, we collect online support images by Google keyword search. We rank the images by visual relevance in descending order (auto-generated by Google), and select the top $N$ supports to conduct the $N$-shot FSS experiment.
Note that the corresponding support masks are generated from the large vision model Grounding-SAM~\cite{ren2401grounded}. We use the support image as the input and its category name as the text prompt, then we directly use the output from Grounding-SAM as our support mask. 
However, as shown in Fig.~\ref{fig: noise}, the predicted masks are always not accurate enough, e.g., the boundary can be coarse and there may exist noisy mask regions (the top row). We thus use a semi-automatic labeling tool~\cite{Wada_Labelme_Image_Polygonal} to efficiently correct these masks (the bottom row). We mark the methods using the refined masks with *.
See Tab.~\ref{tab: llm-comp}, our mask precision is consistently better than DCAMA(Swin-B) and keeps increasing as $N$ gets larger. Besides, we can observe that our method is less sensitive to the noisy support mask. Given the corrected masks, our method* can get slightly better masks, but the outperformance gains relatively smaller margin, showing that our method is more stable against noise.
%
\begin{figure*}
    \centering   \includegraphics[width=1\textwidth]{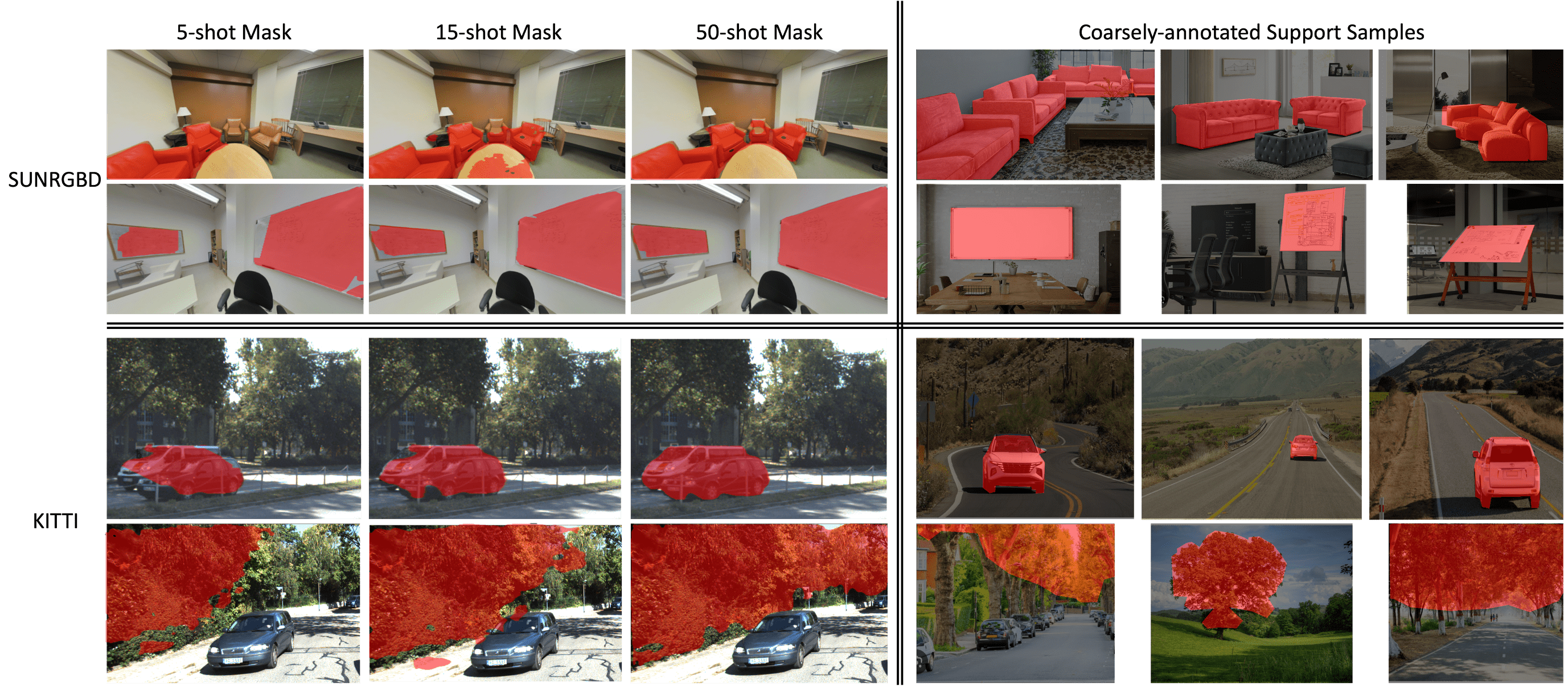}
    \caption{Qualitative results on indoor dataset SUNRGBD (top two rows) and autonomous driving dataset KITTI (bottom two rows). Since our method is robust against support dilution, we can get better segmentation results by simply increasing the support number, see the segmentation improvement from $N=5$ to $N=50$. Due to space limit, we illustrate three support samples for each query, each support is coarsely and quickly annotated with a little human effort.}
    \label{fig: demo single}
\end{figure*}
\begin{table}[t]
\caption{mIoU(\%) results of ablation experiments.}
\label{tab: ablation}
\centering
\resizebox{0.45\textwidth}{!}{
\begin{tabular}{c||cccccc}
\hline
 \multirow{2}{*}{Methods} & \multicolumn{5}{c}{shots}
        \\ \cline{2-7}
      & \multicolumn{1}{c|}{1}     & \multicolumn{1}{c|}{5}     & \multicolumn{1}{c|}{10}    & \multicolumn{1}{c|}{30}    & \multicolumn{1}{c|}{50} & \multicolumn{1}{c}{70}   \\ \hline
      Baseline  & \multicolumn{1}{c|}{41.2} & \multicolumn{1}{c|}{44.7} & \multicolumn{1}{c|}{45.0} & \multicolumn{1}{c|}{44.3} &   \multicolumn{1}{c|}{42.9}& 41.9  \\ \hline
+SC  & \multicolumn{1}{c|}{\textbf{41.5}} & \multicolumn{1}{c|}{\textbf{50.6}} & \multicolumn{1}{c|}{\textbf{53.1}} & \multicolumn{1}{c|}{\textbf{55.4}} &   \multicolumn{1}{c|}{54.2}&55.7  \\ \hline
+SC +Pruning  & \multicolumn{1}{c|}{-} & \multicolumn{1}{c|}{-} & \multicolumn{1}{c|}{-} & \multicolumn{1}{c|}{-} &  \multicolumn{1}{c|}{\textbf{55.5}} &  \textbf{56.1}   \\ \hline
\end{tabular}}
\end{table}
\begin{table}[t]
\caption{mIoU(\%) results of using different support retrieval strategies.}
\label{tab: dis}
\centering
\resizebox{0.45\textwidth}{!}
{
\begin{tabular}{c||cccc}
\hline
 \multirow{2}{*}{Methods} & \multicolumn{4}{c}{shots}
        \\ \cline{2-5}
      & \multicolumn{1}{c|}{30}     & \multicolumn{1}{c|}{50}     & \multicolumn{1}{c|}{70}   & \multicolumn{1}{c}{100}   \\ \hline
Enc-L2 Dis  & \multicolumn{1}{c|}{55.4} & \multicolumn{1}{c|}{54.0} & \multicolumn{1}{c|}{53.6} & 55.1     \\ \hline
Enc-Cos Dis  & \multicolumn{1}{c|}{55.4} & \multicolumn{1}{c|}{54.8} & \multicolumn{1}{c|}{55.8} & 52.7     \\ \hline
VGG-Cos Dis  & \multicolumn{1}{c|}{55.4} & \multicolumn{1}{c|}{55.1} &   \multicolumn{1}{c|}{55.3} & 55.9   \\ \hline
Dinov2-Cos Dis  & \multicolumn{1}{c|}{55.4} & \multicolumn{1}{c|}{\textbf{55.7}} &   \multicolumn{1}{c|}{55.8}  & 56.2 \\ \hline
Support Image Pruning  & \multicolumn{1}{c|}{55.4} & \multicolumn{1}{c|}{55.5} &   \multicolumn{1}{c|}{\textbf{56.1}} & \textbf{56.5}    \\ \hline
\end{tabular}}
\end{table}
\begin{table}[t]
\caption{Test-time analysis of time and memory costs.}
\centering
\label{tab: costs}
\resizebox{0.5\textwidth}{!}
{
\begin{tabular}{c||cccc}
\hline
 \multirow{2}{*}{time(ms) / memory(GB)} & \multicolumn{4}{c}{shots}
        \\ \cline{2-5}
& \multicolumn{1}{c|}{30}     & \multicolumn{1}{c|}{50}     & \multicolumn{1}{c|}{70}  &100    \\ \hline
MSANet  & \multicolumn{1}{c|}{48.6/23.5} & \multicolumn{1}{c|}{71.2/45.3} & \multicolumn{1}{c|}{98.4/68.9} & 130.5/90.1    \\ \hline
DCAMA  & \multicolumn{1}{c|}{26.3/19.3} & \multicolumn{1}{c|}{47.6/26.8} & \multicolumn{1}{c|}{62.8/35.6} & 94.3/45.2     \\ \hline
Ours(w/o Pruning)  & \multicolumn{1}{c|}{26.3/19.3} & \multicolumn{1}{c|}{47.6/26.8} & \multicolumn{1}{c|}{62.8/35.6} & 94.3/45.2     \\ \hline
Ours(w/ Pruning) &\multicolumn{1}{c|}{26.3/19.3} & \multicolumn{1}{c|}{\textbf{30.8}/\textbf{19.3}} & \multicolumn{1}{c|}{\textbf{32.4}/\textbf{19.3}} & \textbf{33.6}/\textbf{19.3}      \\ \hline
\end{tabular}}
\end{table}
%
%
\subsection{Ablation Studies and Test-time Analysis}
\label{exp: ablation}
In Tab.~\ref{tab: ablation}, we ablate the major components of our framework(Swin-B) on COCO-20$^i$ and report the average results, we observe that both SC and Support Image Pruning can contribute to improving the mask mIoU, and our full pipeline attains the highest ratings. Especially when $N$ gets very large (e.g., 70), SC shows significant efforts in enhancing the network against support dilution. We apply the pruning operation only given more than 30 shots, so we leave several blank entries in Tab.~\ref{tab: ablation}.

We also compare the effectiveness of Support Image Pruning with other pre-processing image retrieval techniques. 
In Tab.~\ref{tab: dis}, `Enc-L2 Dis' means we use the Euclidean distance between the encoded support feature and the encoded query feature to select $N'$ relevant supports from $N$ supports; `Enc-Cos Dis' means we use the Cosine distance between encoded features; `VGG-Cos Dis' and `Dinov2-Dis Cos' means we use VGG-19~\cite{simonyan2014very} or the foundation model Dinov2~\cite{oquab2023dinov2} to extract deep features then use Cosine distance for image filtering. Quantitative results verify the superiority of our Pruning module. 

In Tab.~\ref{tab: costs}, we report the per-image inference times and the memory costs. Comparison results with other FSS methods demonstrate that, when $N$ gets larger, our method can achieve higher segmentation performance with comparable or even lower computational costs.
\subsection{Real-world Demonstrations}
\label{exp: real-world}
\begin{figure}
    \centering   \includegraphics[width=0.5\textwidth]{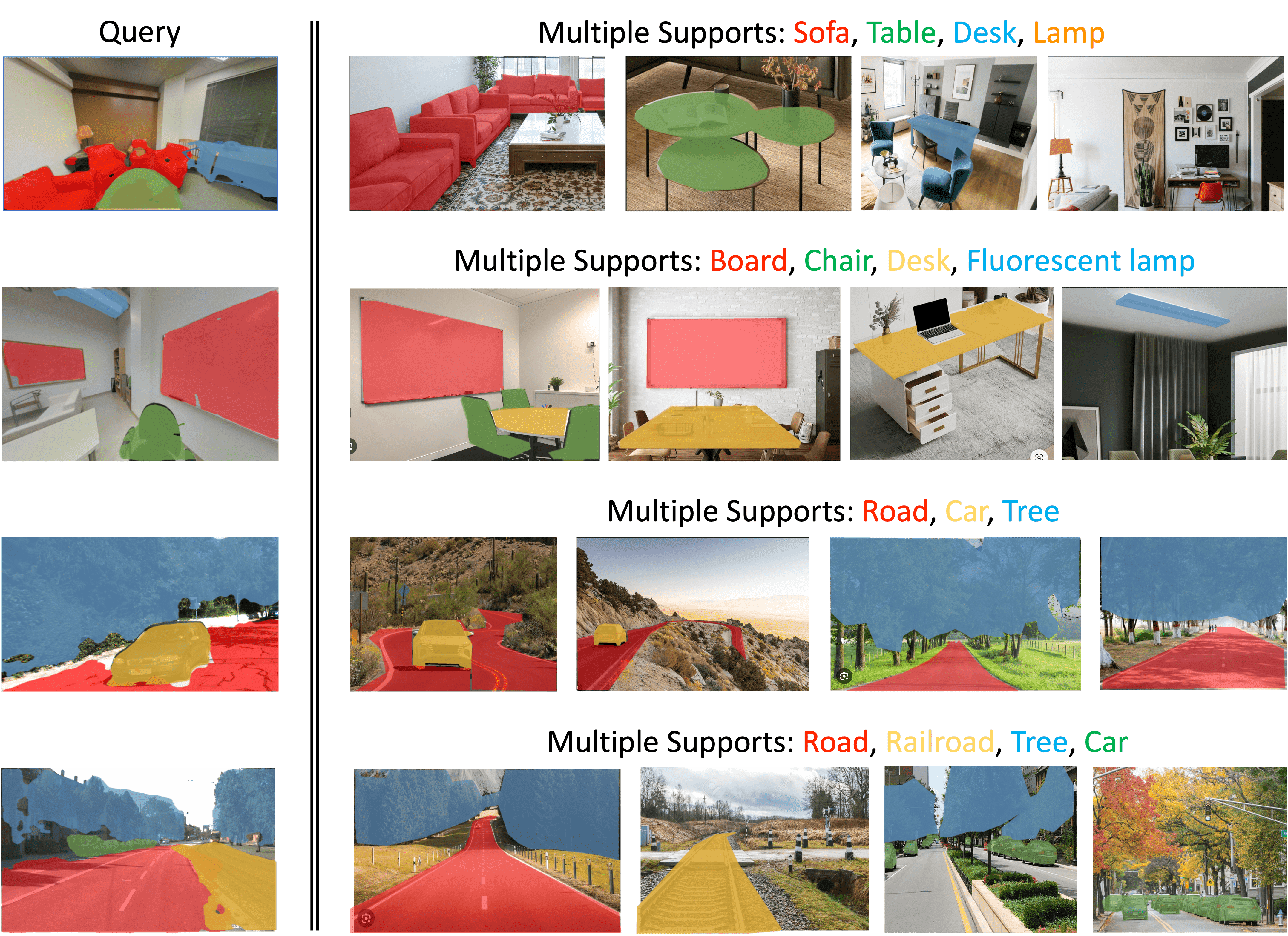}
    \caption{Qualitative results for complex (i.e., multi-category) scenarios of SUNRGBD (top two rows) and KITTI (bottom two rows). In each support set, we mark different categories with different colors. We combine per-category results as the final prediction.}
    \label{fig: demo multi}
\end{figure}
More than adapting basic FSS benchmarks, we want to explore if our method can serve images in real-world tasks, e.g., autonomous driving and indoor navigation.
Therefore, we apply the framework on two 3D benchmarks, SUNRGBD~\cite{song2015sun} and KITTI~\cite{geiger2012we}.
SUNRGBD compiles 47 3D indoor layouts with 800 different object categories to facilitate indoor scene understanding. Given a 3D layout, we extract the RGB image with a random camera view as the 2D query, and collect online images as supports (Sec.~\ref{exp: online}).
KITTI 3D object detection dataset includes 12,000 autonomous driving images of 16 object categories, we pick samples from KITTI Cars Moderate Split as queries, similarly, we use online supports.
Note that our support masks are quickly and coarsely annotated with a little human labor. The model is pretrained on COCO-20$^i$ and directly applied for real-world scenarios, without seeing a single image from SUNRGBD or KITTI.

Typical qualitative results on common objects can be found in Fig.~\ref{fig: demo single}. Our method achieves convincing results for both SUNRGBD (top) and KITTI (bottom), we also observe that given more corresponding supports (i.e., $N$ grows from 5 to 15 to 50), the segmentation masks get more precise and the mask boundaries become smoother.

Considering that real-world scenarios are always composed of objects from various categories, we conduct experiments for those complex cases, shown in Fig.~\ref{fig: demo multi}. We retrieve more informative supports by searching with multiple keywords, then build a 10-shot support set with various categories (we only illustrate four supports in Fig.~\ref{fig: demo multi} for briefness). For each category, we run the model once, and unify the predictions as the final result. 

\begin{figure}
    \centering   \includegraphics[width=0.5\textwidth]{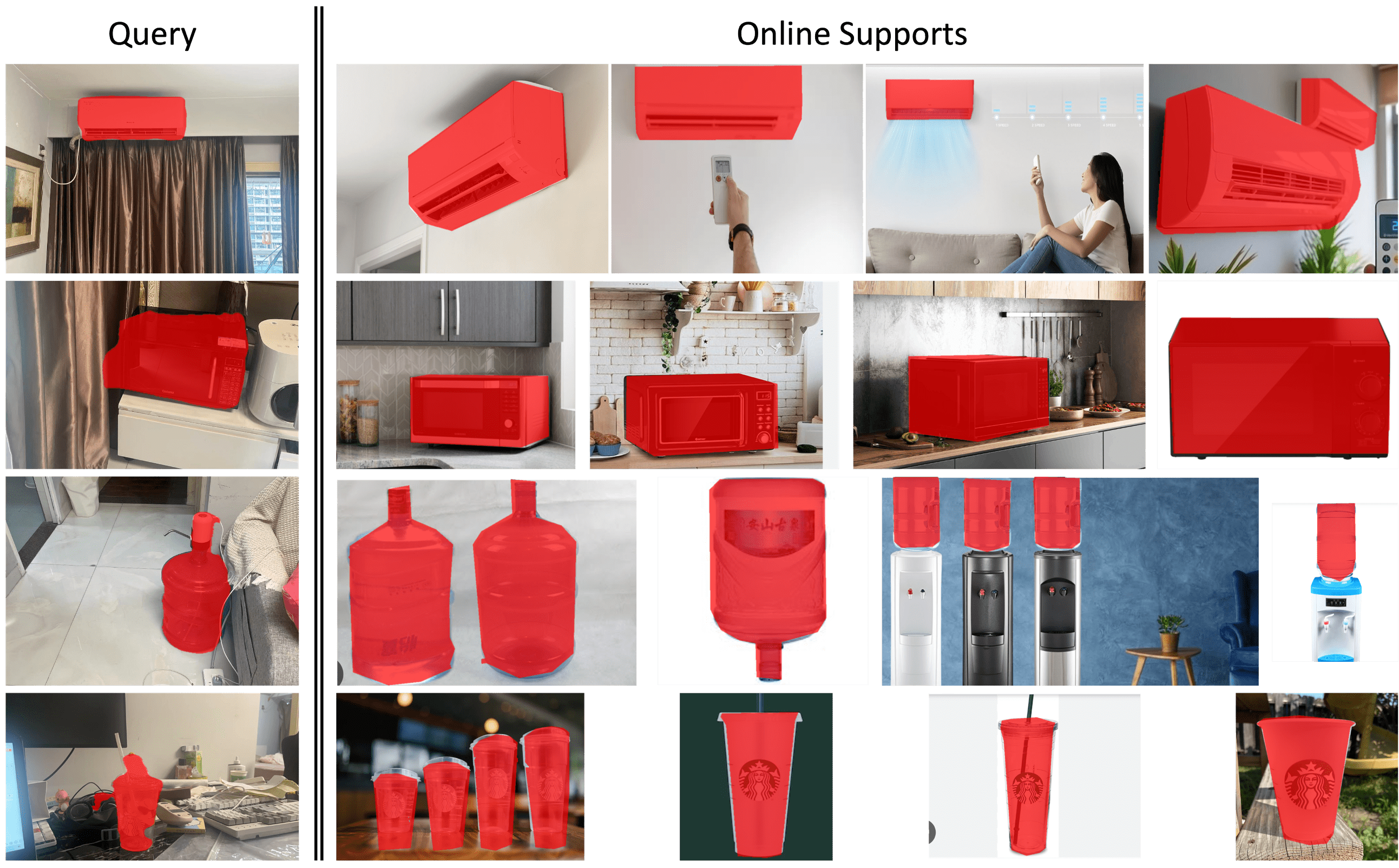}
    \caption{Real-time qualitative results on queries that we randomly captured with an iPhone13. For each query, we use a 15-shot online support set and illustrate only four supports due to space limit. From top to bottom: air conditioner, micro oven, water bucket, translucent Starbucks bottle.}
    \label{fig: demo yes}
\end{figure}

\begin{figure}
    \centering   \includegraphics[width=0.5\textwidth]{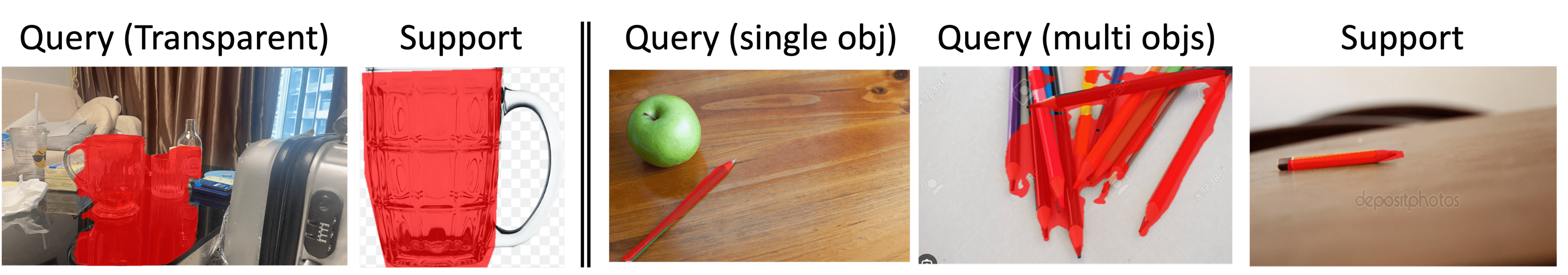}
    \caption{Typical failure cases in 1-shot testing. Left: the model outputs false positive predictions for the reflection regions of the transparent objects; right: mask quality becomes worse when increasing the instance number, the model outputs noisy masks for objects cluttered with occlusion.}
    \label{fig: demo no}
\end{figure}

More than benchmarked scenarios from SUNRGBD and KITTI, we take a step forward to deploy our method on a mobile phone for real-time real-world inference. We randomly capture the surroundings as queries with an iPhone13 and use online images annotated by category-prompted Grounding-SAM as supports. We integrate the above procedures as a user-friendly demo, and we will release it upon paper acceptance. In Fig.~\ref{fig: demo yes}, we show considerable good results on various daily objects. However, in Fig.~\ref{fig: demo no}, we find two types of failure cases. First, the model can generate false positive predictions for irregular objects (e.g., transparent objects); second, when the scenario gets more cluttering, the segmentation performance simultaneously gets worse. These FSS phenomena provide potential research orientations in our future works.

\section{Conclusion}
In this work, we focus on the support dilution problem in Few-shot Semantic Segmentation (FSS). For previous FSS approaches, we find that given more support images (i.e., increasing the shot number), the segmentation performance has little improvement or even goes down. The reason is that, a big support pool includes lots of low-contributed supports holding little or even negative guidance to the query, hence the informative (i.e., high-contributed) supports are diluted and lose their power in support-query correlation learning.
We propose a robust framework against support dilution. First, we design a contribution index to quantitatively measure the true contribution of each support, such that we can know if a high-contributed support dilutes and how bad it dilutes. Based on this prior knowledge, we design a Symmetric Correlation (SC), it can preserve and enhance the high-contributed supports, meanwhile suppress the low-contributed supports. Finally, we develop the Support Image Pruning operation, where we retrieve a compact subset from the big support set, such that SC
can pay less computation effort to concentrate only on relevant supports instead of dealing all of them. We conduct extensive benchmark experiments, where our framework significantly outperforms previous FSS approaches. We also present lots of interesting real-world demonstrations, showing that our method has strong potential for practical usage.

\bibliographystyle{splncs04}
\bibliography{egbib}

\end{document}